\documentclass[11pt]{article}

\usepackage[final]{acl}

\usepackage{times}
\usepackage{latexsym}
\usepackage[T1]{fontenc}
\usepackage[utf8]{inputenc}

\usepackage{microtype}

\usepackage{inconsolata}

\usepackage{graphicx}
\usepackage{array}
\usepackage{booktabs}
\usepackage{amsmath}
\usepackage{enumitem}
\usepackage{url}
\usepackage[many]{tcolorbox}
\tcbuselibrary{breakable}
\usepackage{longtable}
\usepackage{subcaption}

\title{MUCnoHARM@GermEval Shared Task 2026: Retrieval-based In-Context Learning for Defamatory Offences, and Where It Falls Short}

  \author{
 \textbf{Kristin Gnadt\textsuperscript{1,2}},
 \textbf{Maximilian Meidinger\textsuperscript{1}},
 \textbf{Matthias Aßenmacher\textsuperscript{2,3}}
\\
\\
 \textsuperscript{1}Central Office for Information Technology in the Security Sector (ZITiS),  Munich, Germany,\\
 \textsuperscript{2}Department of Statistics, LMU Munich, Germany,\\
 \textsuperscript{3}Munich Center for Machine Learning (MCML), Germany
\\
 \small{
   \textbf{Correspondence:} \href{mailto:kristin.gnadt@zitis.bund.de}{kristin.gnadt@zitis.bund.de}
 }
}

\begin{document}
\maketitle
\begin{abstract} With hate speech being ubiquitous online, automatic detection is crucial, in particular when it comes to criminally relevant social media posts. We study a variety of retrieval-based in-context learning (RetICL) strategies for detecting defamatory offences under \mbox{§§\,185--187 StGB} (the subject of GermEval 2026 Subtask 4).
Few-shot prompting beats zero-shot, but retrieval-based approaches offer only marginal gains over random demonstrations, and even fall behind an optimised static set of demonstrations.
Providing concrete legal knowledge helps, yet model choice outweighs every other system choice. Models over-predict criminal relevance while still missing 26--57\% of criminally relevant posts, suiting them for triage rather than autonomous moderation.

\end{abstract}

\section{Introduction}

Hate speech detection has become a socially consequential application domain for LLM-based text classification. Hate speech is highly prevalent online: In Germany, 34\% of internet users reported encountering instances thereof in the first quarter of 2025 \citep{destatis2025hatespeech}. The volume of potentially harmful content has far outstripped the capacity of human moderation, making automated detection a critical concern for platforms and policymakers alike \citep{european2023online}.
Large language models (LLMs) have demonstrated strong capabilities in addressing the core semantic challenges of automated hate speech detection \citep{albladi2025hate, kums-etal-2025-novel}, making them a natural fit for the task.
To date, most research has focused on English-language datasets and settings \citep{albladi2025hate, usman2025large, gandhi2024hate}, although a growing body of work addresses hate speech detection in other languages, including German \citep{goldzycher-etal-2024-improving, glasebach_gmhp7k_2024, demus_detox_2022}.

Detecting \emph{criminally relevant} hate speech, however, introduces a further layer of complexity:
models must be sensitive to the specific thresholds and distinctions encoded in (country-specific) law, a requirement that current systems only partially meet \citep{schafer2023bias, kums-etal-2025-novel, ludwig_conditioning_2025}. Subtask~4 (DEF) of the GermEval Shared Task 2026 on \emph{Harmful Content Detection in Social Media} instantiates precisely this challenge: classifying posts according to whether they constitute defamatory offences under \mbox{§§\,185--187 StGB} (German Criminal Code) \citep{felser2026germeval}.

\begin{figure}[t]
    \centering
    \includegraphics[width=0.95\linewidth]{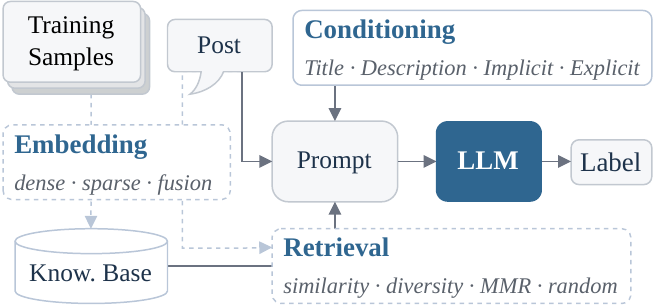}
    \caption{RetICL framework. Embedded training samples make up a knowledge base, from which few-shot demonstrations are retrieved, which are given to a model along with a task description (\emph{conditioning}) and the post.}
    \label{fig:framework}
\end{figure}

\paragraph{Contributions.} In this work, we test different prompting strategies for the detection of defamatory offences. Our RetICL framework consists of two main components (Fig.~\ref{fig:framework}):

\begin{enumerate}[noitemsep]
    \item The legal information on \mbox{§§\,185--187 StGB} %and the scheme by \citet{zufall-etal-2019-legal} used to annotate the dataset
    that is provided to the model (\textbf{Conditioning}); and
    \item The retrieval mechanisms for few-shot demonstrations (\textbf{Embedding} and \textbf{Retrieval}).
\end{enumerate}

The first component builds on \citet{ludwig_conditioning_2025}, who use prompts of differing abstraction levels---``conditioning'' models on different degrees of legal knowledge---to detect whether a post falls under §\,130 StGB (incitement to hatred) (Section~\ref{subsec:legal-conditioning}). 
The second component tests different strategies for selecting demonstrations for few-shot prompting, which are also compared to zero-shot prompting.
%We compare zero-shot prompting and several retrieval-based strategies.
%, and---as an ablation---an optimised static set of demonstrations.
%After presenting relevant related work in Section~\ref{sec:background}, we explain our methodology in Section~\ref{sec:methods} as well as experimental settings in Section~\ref{sec:exp}. Results are presented in Section~\ref{sec:res} and discussed in Section~\ref{sec:discussion}. 
Section~\ref{sec:background} situates our two components in prior work on legal conditioning and RetICL. Section~\ref{sec:methods} details the conditioning modes and retrieval strategies we compare, and Section~\ref{sec:exp} the experimental setups. Section~\ref{sec:res} reports that concrete legal knowledge and few-shot prompting help, but retrieval strategy barely matters, with the full discussion of implications in Section~\ref{sec:discussion}.

\section{Background \& Related Work}\label{sec:background}

\subsection{Defamatory Offences}\label{subsec:annotation-scheme}

The StGB sections central to this task are the three defamatory offences in \mbox{§§\,185--187}, which all protect a person's right to honour and differ mainly in whether the disparaging statement is an opinion or a factual claim, and to whom it is addressed \citep{bundestag2022}: \mbox{§\,185} \emph{Insult} (\emph{Beleidigung}), \mbox{§\,186} \emph{Defamation} (\emph{Üble Nachrede}), and \mbox{§\,187} \emph{Intentional Defamation} (\emph{Verleumdung}).

The dataset (GermEval 2026, Subtask~4)
%\footnote{\url{https://www.codabench.org/competitions/14006/}}
was annotated with the decision scheme of \citet{zufall-etal-2019-legal}. Derived directly from the statutory norms in the German deductive civil-law tradition, the scheme operationalises the legal assessment of \mbox{§§\,185--187} StGB as a sequence of six binary (yes~/~no) decisions based solely on the text of a post. These decisions cover whether the post targets a valid holder of the right to honour, whether it is disparaging, whether it is a factual claim or a value judgment, and how freedom of expression is weighed against the right to honour (§\,193 StGB). The full scheme is provided in Appendix~\ref{app:annotation-scheme}.
%We provide the full scheme in Appendix~\ref{app:annotation-scheme}.

Because it decomposes a complex legal judgment into independently checkable sub-decisions, the scheme yields reliable annotations---\citet{zufall-etal-2019-legal} report that instructed laypeople apply it with reasonable reliability against an expert reference. Additionally, the scheme provides a natural source of graded legal knowledge for prompting, which our \emph{conditioning} settings utilise (Section~\ref{sec:methods}).

\subsection{Legal Conditioning}
\label{subsec:legal-conditioning}

How a task is presented to an LLM strongly affects its performance. Even meaning-preserving changes to a prompt can shift performance substantially \citep{sclar2024quantifying,gan2023sensitivityrobustnesslargelanguage,lu-etal-2022-fantastically}. Beyond such surface-level choices, the \emph{content} of the prompt---which task-relevant knowledge it makes available to the model---is itself a design decision. This is especially pertinent in specialised domains such as law, where general prompting techniques have proven helpful but require careful adaptation \citep{trautmann2022legal, HakimiParizi2023ACS}, raising the question of \emph{how much} and \emph{what kind} of legal knowledge to place in the prompt.

\citet{ludwig_conditioning_2025} address this question for hate speech falling under §\,130 StGB (incitement to hatred) by \emph{conditioning} LLMs on legal knowledge at different abstraction levels. Their prompts range from the title of a norm---relying on whatever knowledge the model has internalised---
to the verbatim or simplified statutory text, and finally to an explicit decomposition of the offence into its constituent subtasks.
The decomposition follows the subtask-based framing that \citet{alkhamissi2022token} found effective for few-shot hate speech detection. 
Following these approaches, we vary the legal knowledge supplied to the model, from merely stating the relevant section numbers and titles to decomposing the task into the subtasks of the annotation scheme of \citet{zufall-etal-2019-legal}.

Counterintuitively, \citet{ludwig_conditioning_2025} found that models conditioned on more abstract knowledge outperformed those given more concrete legal information. A substantial performance gap remained between LLM prompting approaches and legal experts, and the models could barely match laypeople applying the annotation scheme \citep{ludwig_conditioning_2025, zufall-etal-2022-legal}. Conversely, \citet{voncossel2026imperfectalternativesrulemappingneurosymbolic} found that subtask decomposition for detecting §\,130 StGB offences, combined with a logical scaffold in a neuro-symbolic approach, outperformed less complex prompting approaches.

\subsection{Retrieval-based In-Context Learning}

LLMs can perform classification through in-context learning (ICL), conditioning on a few input-output demonstrations in the prompt without updating model parameters. ICL, however, is highly sensitive to the choice, number, ordering, and format of those demonstrations \citep{luo2024context}. A fixed, manually curated, or randomly sampled set applied uniformly to every input is therefore suboptimal, which motivates RetICL\footnote{In the literature, RetICL is also referred to as dynamic few-shot prompting, RAG-enhanced LLMs, (selective) demonstration retrieval, and active learning for ICL.}, in which demonstrations are retrieved per query from a labelled pool \citep{luo2024context}.

Retrieval is usually guided by a similarity objective, most commonly selecting the top-$k$ examples by using sparse term-matching retrievers such as BM25 \citep{robertson-1976-relevance,robertson_bm25_2009} or dense sentence-embedding retrievers scored by cosine similarity \citep{luo2024context, fan-2024-survey}. \citet{margatina-etal-2023-active} show that similarity-based retrieval consistently outperforms uncertainty-, diversity-, and random-based selection, while zero-shot prompting performs worst for text classification tasks. 
\citet{miller_dynamic_2025} find RetICL best across seven open-weights LLMs on clinical note section classification, raising $F1_{\text{macro}}$ substantially over both zero-shot and static few-shot; they note that static few-shot is sometimes no better than zero-shot, that smaller models benefit more, and that more demonstrations do not monotonically help.

These methods are increasingly applied to abusive language and content moderation tasks adjacent to hate speech detection. For implicit hate speech, \citet{kim_selective_2025} prioritise demonstrations sharing the target group before falling back to BM25 similarity, reducing the over-sensitivity of LLMs to toxic surface terms. Related work uses retrieval to discover emergent dog whistles \citep[coded language used to evade automatic detection mechanisms;][]{sasse-etal-2025-making}. The evidence is not uniformly positive, however: \citet{liu2024poliprompt} found dynamic exemplar selection less reliable than their prompt-optimisation framework.

RetICL is a strong but task-dependent strategy, yet the relative behaviour of zero-shot, static, and RetICL remains unexamined for \emph{criminally relevant} hate speech, where classification depends on a precise legal threshold rather than a general notion of toxicity. We address this gap by systematically testing retrieval mechanisms and prompting strategies for detecting defamatory offences as part of the shared task.

\section{Materials and Methods} \label{sec:methods}

\paragraph{Dataset.}
The DEF task is based on a corpus of German-language tweets, each labelled according to the annotation scheme of \citet{zufall-etal-2019-legal} described in Section~\ref{subsec:annotation-scheme}, with a binary label indicating whether the post falls under \mbox{§§\,185--187 StGB}. The labelled portion comprises $3{,}263$ tweets, on which we report cross-validated results; a further $577$ tweets form the unlabelled competition test set, which is used only for the shared task submission. In the training dataset, ${\sim}13\%$ of the samples are labelled as positive (criminally relevant).
In addition, one author annotated a subset of $797$ training tweets at the level of each individual decision step of the scheme according to the explanations of \citet{zufall-etal-2019-legal}. Whereas the organisers' data provide only the final label, these auxiliary annotations record the outcome of every step; they are used as the demonstration pool for few-shot retrieval when deconstructing the classification task into individual steps (\emph{Explicit} conditioning setting).

\paragraph{Models.}
Four instruction-tuned models spanning different families, parameter scales, and degrees of openness are evaluated: \mbox{Gemma-4 26B} (\texttt{gemma-4-26B-A4B-it}), \mbox{Gemma-4 E4B} (\texttt{gemma-4-E4B-it}), \mbox{Qwen3.5 9B} (\texttt{Qwen3.5-9B}), and \mbox{EuroLLM 22B} (\texttt{EuroLLM-22B-Instruct-2512}). All models are loaded from the HuggingFace Hub and run with the Transformers library \citep{wolf-etal-2020-transformers}.

\paragraph{Conditioning.}
The first component varies the legal information supplied to the model, following \citet{ludwig_conditioning_2025}.
Each \emph{conditioning} mode corresponds to a different task description. The \emph{Title} setting states only the numbers and titles of the relevant StGB sections. The \emph{Description} setting additionally provides a description of what constitutes a defamatory offence under the annotation scheme (Section~\ref{subsec:annotation-scheme}). The \emph{Implicit} setting presents all six decision steps of the scheme inline, so that the model traverses the full decision tree within a single inference call. The \emph{Explicit} setting instead decomposes the task into the six steps as separate inference calls: each step is classified on its own, and the final label is derived from the sequence of step outcomes. Each step is run as an independent prompt containing only that step's instructions, its demonstrations, and the post. All prompts and additional information on chat template and few-shot integration are provided in Appendix~\ref{app:prompts}.

\paragraph{Demonstration selection.}
The second component governs how few-shot demonstrations are chosen. In the \emph{zero-shot} configuration ($k=0$), no demonstrations are included. In the \emph{few-shot} configuration, we include $k$ demonstrations. 
The class \emph{ratio} of demonstrations and the within-prompt \emph{ordering} of demonstrations are varied for the ablation study (Section~\ref{ablation}).
Demonstrations are selected either \emph{statically}---a single fixed set, chosen once and reused for every test instance---or \emph{dynamically}, where a separate set is retrieved per instance at inference time. The dynamic strategies are our main object of study and are defined by the embedding and retrieval modes below.

\paragraph{Embedding mode.}
Dynamic retrieval draws demonstrations from a knowledge base built over the training split and stored in one of three ways. In the \emph{dense} setting, each example is embedded with \texttt{codefuse-ai/F2LLM-v2-1.7B} \citep{f2llm-v2}, the highest-ranked model in its size range on the MTEB leaderboard for German tasks \citep{muennighoff-etal-2023-mteb} throughout the entire duration of this project in the first half of 2026 (see Appendix~\ref{app:knowledge-bases}). The embeddings are stored in a LangChain \texttt{InMemoryVectorStore} \citep{LangChain}. The \emph{sparse} setting uses a LangChain \texttt{BM25Retriever} which vectorises training samples sparsely and retrieves demonstrations based on keyword matches using the BM25 algorithm. The \emph{fusion} setting interleaves the results of both the dense and sparse retrievers.

\paragraph{Retrieval mode.}
Given a knowledge base made up of an embedded training dataset, demonstrations are selected by one of four strategies. \emph{Similarity} retrieval returns the top-$k$ examples by cosine similarity (dense) or BM25 score (sparse). \emph{Diversity} retrieval clusters the dense training embeddings with $k$-means and draws one example at random from each cluster. \emph{MMR} (Maximal Marginal Relevance) re-ranks dense candidates to balance \textit{similarity} against \textit{diversity}. \emph{Random} retrieval samples uniformly from the training pool and serves as the baseline retrieval strategy. Because \textit{MMR} and \textit{diversity} rely on \textit{dense} embeddings, they are applied only in the \textit{dense} embedding setting; \textit{sparse} and \textit{fusion} embeddings are used only for \textit{similarity}-based retrieval.

\paragraph{Baseline.}
As a non-neural reference point, we implement a naive $1$-nearest-neighbour classifier using TF-IDF representations of the training split to assign each test post the label of its single closest training neighbour. It uses no legal domain knowledge and relies entirely on surface-level lexical overlap; conceptually, it corresponds to dynamic few-shot retrieval at $k=1$.

\section{Experiments}\label{sec:exp}

\paragraph{General Setup.} All reported results use stratified four-fold cross-validation over the labelled set: in each fold, three parts form the knowledge base and the held-out part is used for evaluation. The primary metric is $F1_{\text{macro}}$ (unweighted mean of per-class $F1$ scores); we additionally report True-class Precision (P$_\mathsf{T}$) and Recall (R$_\mathsf{T}$). Instances on which a model abstains (no parseable label, model refusal) are excluded from scoring, but abstentions are reported separately. More details on experiment settings are reported in Appendix~\ref{app:detailed-settings}.\footnote{Code Repository: \url{https://github.com/akristing22/GermEval2026-def-prompting}}

\subsection{Exploration}

In order to control the computational cost during the main experiments, we aim to find the optimal number of demonstrations (in terms of $F1_{\text{macro}}$ score) in an exploratory analysis: we test demonstration sizes $k \in \{4,8,16,32\}$ across two prompting configurations and all models.

\subsection{Main Experiments}\label{main_exp}

Given that RetICL is the focus of this research, the main experiments investigate the different strategies across the four open-weights instruction-tuned models \mbox{Gemma-4 26B}, \mbox{Gemma-4 E4B}, \mbox{Qwen3.5 9B}, and \mbox{EuroLLM 22B}, and all four conditioning modes (\emph{Title}, \emph{Description}, \emph{Implicit}, and \emph{Explicit}). For RetICL, we evaluate \emph{random}, \emph{diversity}-, \emph{MMR}-, and \emph{similarity}-based retrieval; the latter for the three different embedding methods \emph{dense}, \emph{sparse}, and \emph{fusion}. Additionally, zero-shot prompting is tested. 
The retrieved demonstrations are \emph{balanced} in terms of their class labels, and their order in the prompt is \emph{random}.

\subsection{Ablations}

\paragraph{Proprietary Model.}

In order to understand how the performance on this complex task scales with model size far beyond the 4B--26B parameter range, we evaluate the latest OpenAI model of the GPT series (GPT-5.5).
Evaluations are run with the \emph{Title} zero-shot setting and with \emph{Explicit} mode with \emph{dense} \emph{similarity}-based demonstration retrieval. 

\paragraph{Fine-Tuning.}

ICL strategies offer lower computational cost and greater adaptability than supervised fine-tuning (SFT) methods. In order to understand how the performance of our ICL strategies compares to an SFT baseline, we fine-tune the smaller \mbox{Gemma-4 E4B} model using the parameter-efficient method QLoRA \citep{qlora}. The model is trained with the \emph{Implicit} conditioning mode in a zero-shot set-up.

\paragraph{Class Ratio and Ordering.}

The main experiments use only \emph{balanced} (w.r.t. class label), \emph{randomly} ordered few-shot demonstrations. We investigate the effects of class ratio and ordering on classification performance by also arranging the demonstrations by class label (\emph{true-first} and \emph{true-last}) and by selecting examples with a class ratio \emph{proportional} to the ratio in the training dataset. These effects are tested for \emph{Implicit} conditioning with \emph{dense}, \emph{similarity}-based retrieval.

\paragraph{Static Demonstrations.}

In order to compare RetICL demonstration approaches not only to zero-shot prompts but to a static few-shot strategy, a set of few-shot demonstrations is tested, which is selected by searching randomly for the best-performing set of demonstrations (more details are reported in Appendix~\ref{app:random-search}).
Two sets of eight demonstrations for \mbox{Gemma-4 26B} are selected, using the \emph{Implicit} and \emph{Title} conditioning modes.

\subsection{Shared Task}

We select five model~/~configuration combinations for running on the held-out competition test set. The best two runs are mentioned in the respective experiment sections; more details are reported in Appendix~\ref{app:shared-task}.

\section{Results}\label{sec:res}

\subsection{Exploration}

\begin{figure}[ht]
    \centering
    \includegraphics[width=.85\linewidth]{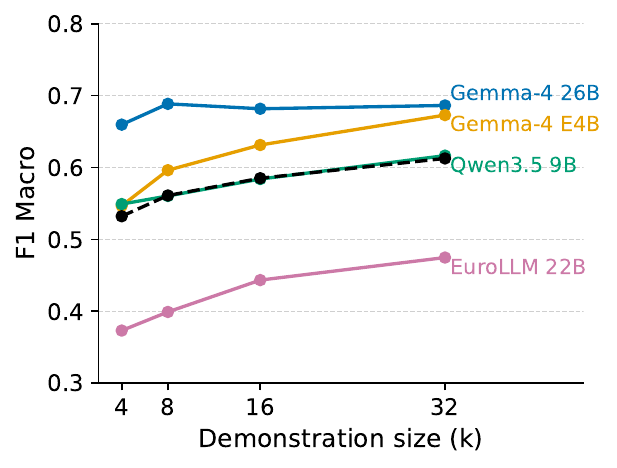}
    \caption{Exploration: Mean $F1_{\text{macro}}$ by demonstration size, per model. The black dotted line shows the mean $F1_{\text{macro}}$ per demonstration size across models.}
    \label{fig:demo_size_exploration}
\end{figure}

Across models, the largest average gain comes from increasing the demonstration count from $4$ to $8$ ($+0.029$). Beyond $k=8$, larger sizes decrease performance slightly for \mbox{Gemma-4 26B}, whereas performance continues to improve for the other three models (Fig.~\ref{fig:demo_size_exploration}). Scores are reported in detail in Appendix~\ref{app:exploration-results}.
To limit computational cost, we set $k=8$ for all subsequent experiments, even though larger sizes would likely benefit the other three models.

\subsection{Main Experiments}

\begin{figure*}[ht]
    \centering
    \includegraphics[width=\textwidth]{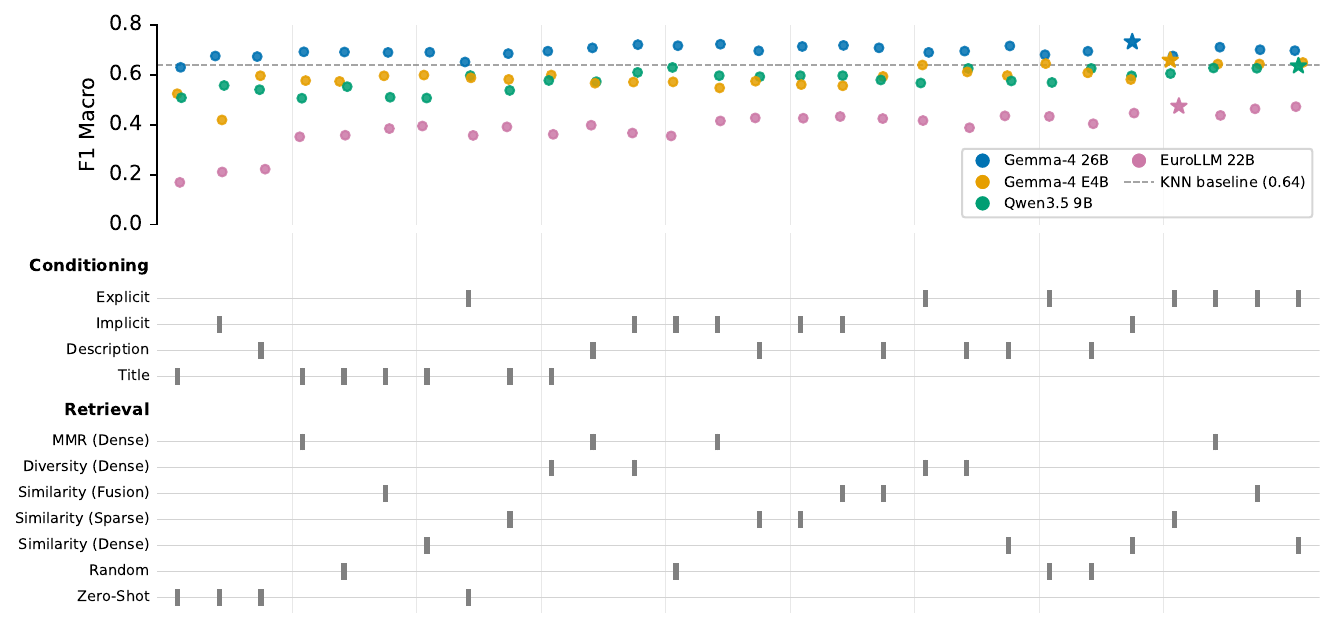}
    \caption{Each point shows the $F1_{\text{macro}}$ score of one model and prompting configuration. Configurations are sorted left to right by increasing cross-model mean $F1_{\text{macro}}$. The lower panel shows which conditioning and retrieval choices correspond to each position on the curve. The best configuration per model is indicated by a star. The dotted line shows the naive nearest-neighbour classifier used as a baseline.}
    \label{fig:specification}
\end{figure*}

A full score table can be found in Appendix~\ref{app:results}.
Figure~\ref{fig:specification} reports $F1_{\text{macro}}$ score across all prompting configurations, for all models. The baseline classifier achieves an $F1_{\text{macro}}$ of $0.644$, which only \mbox{Gemma-4 26B} consistently surpasses. The best performing configuration (\mbox{Gemma-4 26B}, \textit{Implicit} conditioning, \textit{dense}, \textit{similarity}-based retrieval) achieves an $F1_{\text{macro}}$ score of $0.733$ on our test set and $0.72$ on the held-out competition set.
As shown in Table~\ref{tab:e-best-per-model}, all models tend to over-predict the positive class, yielding low positive-class Precision; even so, between $26\%$ and $57\%$ of criminally relevant posts go undetected (positive-class Recall $0.43$--$0.74$). 
While the $1$-NN baseline is stronger than some models in terms of $F1_{\text{macro}}$, it falls substantially behind all models in terms of Recall, except for \mbox{Qwen3.5 9B}, which is only marginally better than the baseline ($+0.067$).

\begin{table}[ht]
\centering
\resizebox{.45\textwidth}{!}{
\begin{tabular}{llrrr}
\toprule
Model & Config & $F1_{\text{macro}}$ & P$_\mathsf{T}$ & R$_\mathsf{T}$ \\
\midrule
Gemma-4 26B & Impl. D-Sim & \textbf{0.733}  & \textbf{0.445} & \textbf{0.740} \\
Gemma-4 E4B & Expl. S-Sim & 0.659 & 0.341 & 0.622 \\
Qwen3.5 9B & Expl. D-Sim & 0.636 & 0.338 & 0.426 \\
EuroLLM 22B & Expl. S-Sim & 0.474 & 0.180 & \textbf{0.740} \\
% Gemma-4 E4B (FT) & Impl. ZS & 0.741 & 0.702 & 0.431 \\
% GPT-5.5 & Expl. D-Sim & 0.716& 0.415 & 0.747 \\
$1$-NN & & 0.644 & 0.391 & 0.359\\
\bottomrule
\end{tabular}
}
\caption{Best configuration per model and naive baseline classifier. \textit{Impl.}~/~\textit{Expl.}: Implicit~/~Explicit conditioning. \textit{D}~/~\textit{S}: dense~/~sparse embeddings. \textit{Sim}: similarity-based retrieval. Best scores per metric are indicated in \textbf{bold}.}
\label{tab:e-best-per-model}
\end{table}

Across models, more detailed conditioning tends to improve performance, with \emph{Explicit} best on average. It is the best mode for \mbox{Gemma-4 E4B}, \mbox{Qwen3.5 9B}, and \mbox{EuroLLM 22B}, while \mbox{Gemma-4 26B} benefits most from \emph{Implicit} conditioning (Table~\ref{tab:conditioning-zs-fs}). Except for \mbox{Gemma-4 E4B}, the least detailed \emph{Title} conditioning performs worst. However, the trend is not strictly monotonic: averaged across models, \emph{Description} ($0.567$) slightly exceeds the more concrete \emph{Implicit} mode ($0.559$). Zero-shot prompting yields the worst scores for every model; the gap to few-shot is most pronounced for \mbox{EuroLLM 22B} and smallest for Qwen3.5 9B.

\begin{table}[ht]
\centering
\resizebox{.48\textwidth}{!}{
\begin{tabular}{l|rrrr|rr}
\toprule
Model & Title & Desc. & Impl. & Expl. & ZS & FS\\
\midrule
Gemma-4 26B & 0.683 & 0.699 & \textbf{0.715} & 0.687 & 0.658 & \textbf{0.702} \\
Gemma-4 E4B & 0.579 & 0.593 & 0.545 & \textbf{0.639} & 0.532 & \textbf{0.598} \\
Qwen3.5 9B & 0.529 & 0.588 & 0.598 & \textbf{0.605} & 0.551 & \textbf{0.585}\\
EuroLLM 22B & 0.345 & 0.386 & 0.380 & \textbf{0.437} & 0.240 & \textbf{0.411}\\
\midrule
\emph{Mean} & \emph{0.534} & \emph{0.567} & \emph{0.559} & \emph{\textbf{0.592}} & \emph{0.495} & \emph{\textbf{0.574}} \\
\bottomrule
\end{tabular}
}
\caption{Mean $F1_{\text{macro}}$ per model and conditioning mode, and per zero-shot (ZS) and few-shot (FS), $k=8$. Best results per model are indicated in \textbf{bold}.}\label{tab:conditioning-zs-fs}
\end{table}

% \begin{table}[ht]
% \centering
% \caption{Zero-shot vs.\ few-shot (k=8). Per model mean $F1_{\text{macro}}$ score over all prompting configurations.} \label{tab:a2-model-shot}
% \resizebox{.3\textwidth}{!}{
% \begin{tabular}{lrr}
% \toprule
% Model & zero-shot & few-shot\\
% \midrule
% Gemma-4 26B & 0.658 & 0.702 \\
% Gemma-4 E4B & 0.532 & 0.598 \\
% Qwen3.5 9B & 0.551 & 0.585 \\
% EuroLLM 22B & 0.240 & 0.411 \\
% \bottomrule
% \end{tabular}
% }
% \end{table}

\begin{figure}[ht]
    \centering
    \includegraphics[width=.85\linewidth]{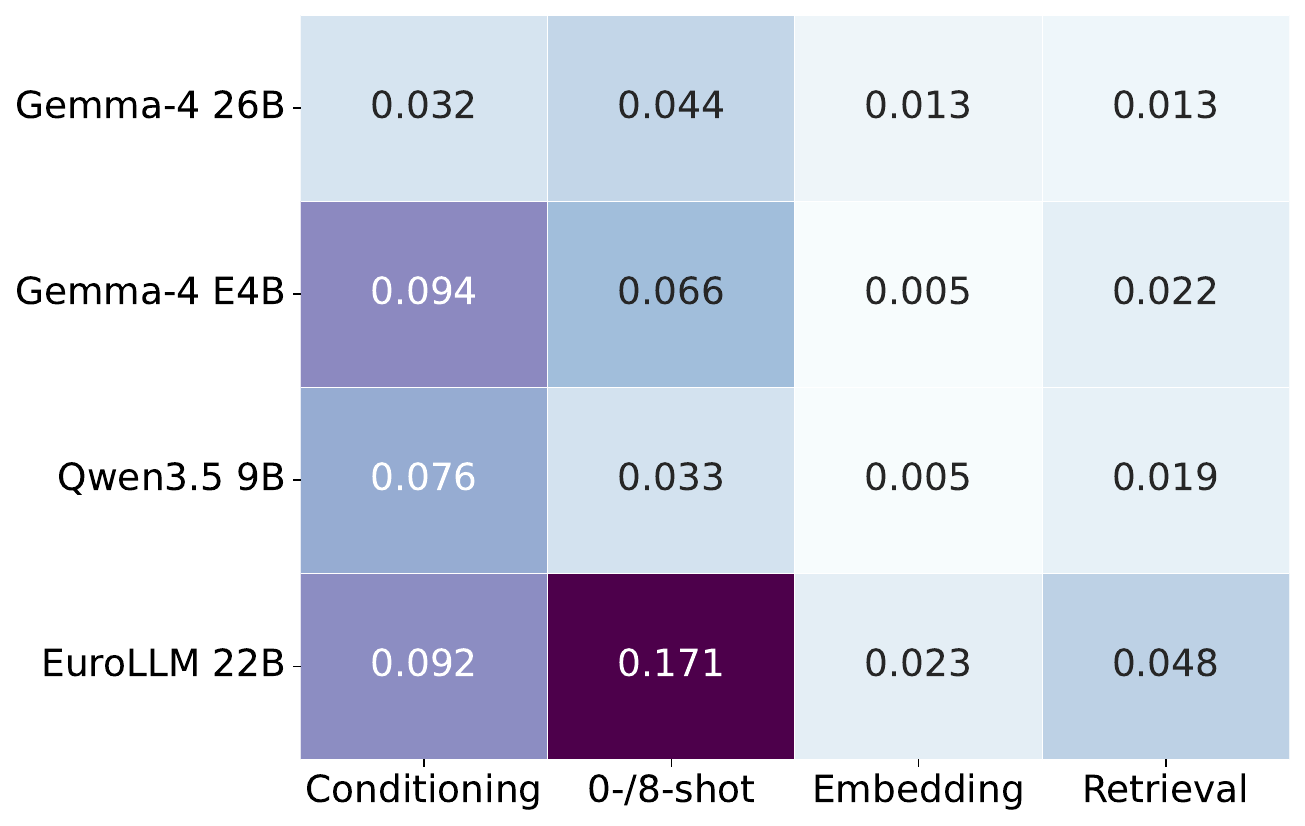}
    \caption{Impact of each configuration axis on $F1_{\text{macro}}$: difference between highest and lowest mean $F1_{\text{macro}}$ across that axis. Each cell reports this range for one model (row) and one configuration dimension (column), marginalising over all other axes.}
    \label{fig:impact_prompting}
\end{figure}

Figure~\ref{fig:impact_prompting} reports the impact of each configuration axis on the performance of each model. Embedding and retrieval choices have little impact on $F1_{\text{macro}}$, whereas the differences across conditioning modes and between zero- and few-shot prompting are more pronounced. The best model (\mbox{Gemma-4 26B}) is least affected by prompting configuration, whereas the weakest (\mbox{EuroLLM 22B}) is most affected, across all axes. For comparison: impact of model choice alone is $0.309$.

\paragraph{Abstentions.}
All scores are computed only on replies that could be parsed into a binary \texttt{True}~/~\texttt{False} label. Across all $28$ prompting configurations per model, fewer than $1\%$ of generations could not be parsed (further details are reported in Appendix~\ref{app:abstentions}).

\subsection{Ablations}\label{ablation}

% \begin{table}[ht]
% \centering
% \resizebox{.45\textwidth}{!}{
% \begin{tabular}{llrrrr}
% \toprule
% Model & Config & $F1_{\text{macro}}$ & P$_\mathsf{T}$ & R$_\mathsf{T}$ \\
% \midrule
% Gemma-4 E4B (FT) & Impl. ZS & 0.741 & 0.702 & 0.431 \\
% GPT-5.5 & Expl. D-Sim & 0.716& 0.415 & 0.747 \\
% GPT-5.5 & Title ZS & 0.440 & 0.192 & 1 \\
% \bottomrule
% \end{tabular}
% }
% \caption{Best configuration per model and naive baseline classifier. \textit{Impl.}/\textit{Expl.}: Implicit/Explicit conditioning. \textit{ZS}: zero-shot. \textit{D-Sim}: dense, similarity-based retreival.}
% \label{tab:ablation-models}
% \end{table}

\paragraph{Proprietary Model.}
Under the minimal \emph{Title} zero-shot setting, GPT-5.5 achieves an $F1_{\text{macro}}$ of $0.440$, beating only \mbox{EuroLLM 22B} in the same configuration and falling substantially behind the other models. With \emph{Explicit} conditioning and \emph{dense}, \emph{similarity}-based retrieval, GPT-5.5 reaches $0.716$, beating the best open-weights model \mbox{Gemma-4 26B} in the same configuration by $0.018$, but falling behind \mbox{Gemma-4 26B}'s best run by $0.017$.

\paragraph{Fine-Tuning.}

Fine-tuned with \emph{Implicit} conditioning in a zero-shot set-up, \mbox{Gemma-4 E4B (FT)} achieves an $F1_{\text{macro}}$ of $0.741$. This is marginally above the best run of the main experiments (\mbox{Gemma-4 26B}, Impl.\ D-Sim; $0.733$; $+0.008$), is a substantial improvement over the best \emph{Implicit} \mbox{Gemma-4 E4B} run ($0.582$; $+0.159$), and is $0.082$ above the best overall \mbox{Gemma-4 E4B} run ($0.659$).

The fine-tuned \mbox{Gemma-4 E4B (FT)} with \textit{Implicit} conditioning also achieves our best results on the held-out competition test set: an $F1_{\text{macro}}$ score of $0.74$. It shows a contrasting pattern in positive-class Precision ($0.702$) and Recall ($0.431$), being less sensitive to defamatory offences but more precise in its predictions.

\paragraph{Class Ratio and Ordering.}

\begin{figure}[ht]
    \centering
    \includegraphics[width=.8\linewidth]{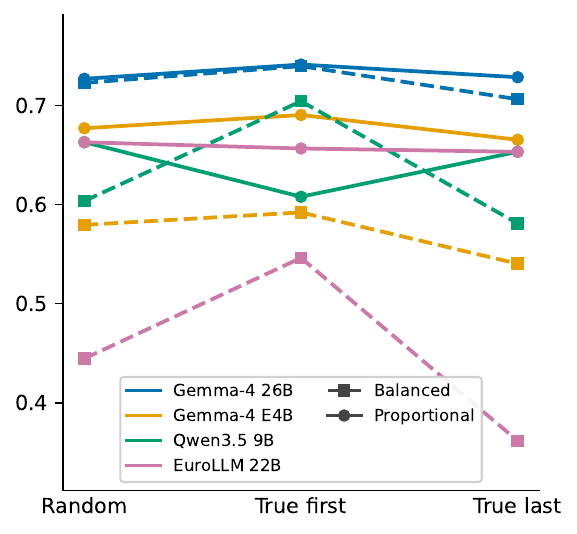}
    \caption{Interaction effects between demonstration \emph{ratio} and \emph{order}. Mean $F1_{\text{macro}}$ scores per model and configuration are shown.}
    \label{fig:interaction_ablations}
\end{figure}

\begin{figure}[ht]
    \centering
    \includegraphics[width=0.75\linewidth]{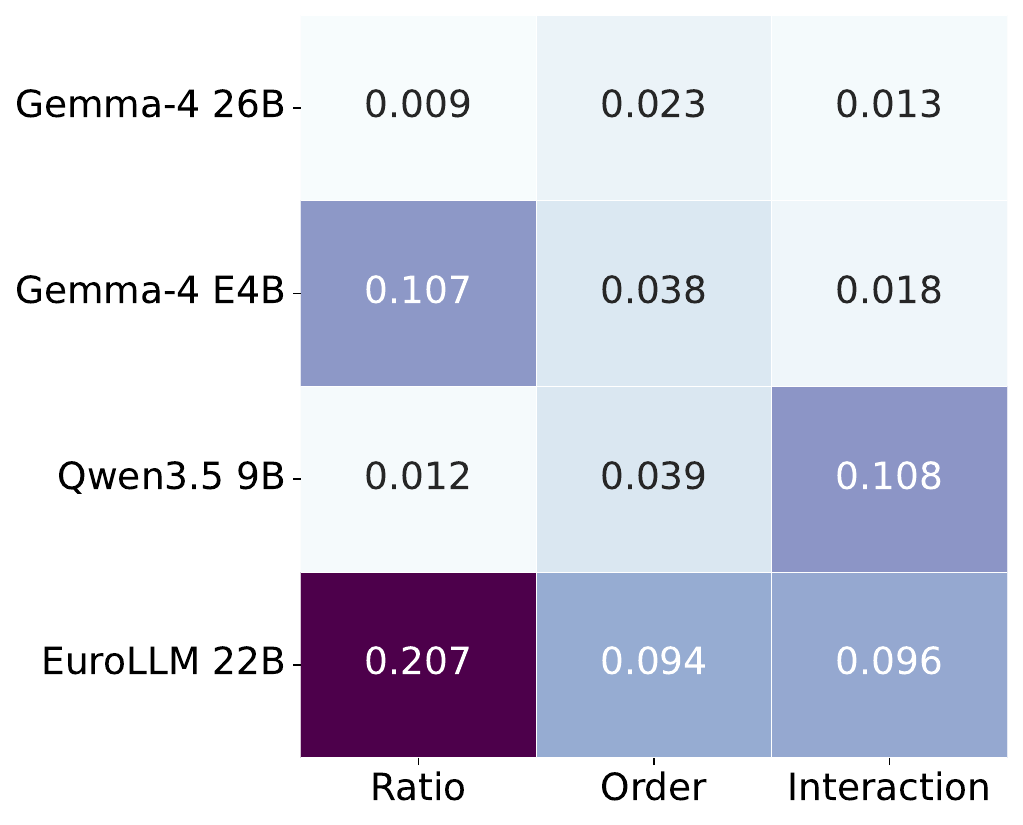}
    \caption{
    Each cell shows the $F1_{\text{macro}}$ impact of one factor for one model. Ratio and Order report the range of marginal means. Interaction reports the range of interaction residuals $\delta(r,o) = \text{F1}(r,o) - \bar{\text{F1}}_{r} - \bar{\text{F1}}_{o} + \bar{\text{F1}}$.
    }
    \label{fig:ablation_heat}
\end{figure}

The impact of demonstration \emph{ratio} and \emph{ordering} (in terms of class labels) on $F1_{\text{macro}}$ is highly model-dependent, as depicted in Figures~\ref{fig:interaction_ablations} and \ref{fig:ablation_heat}. \mbox{Gemma-4 26B} is barely affected. \mbox{Gemma-4 E4B} is only slightly affected by ordering, but the effect of class ratio is more pronounced. \mbox{EuroLLM 22B} is most affected by both class ratio and ordering. The individual ratio and order axes have little effect on \mbox{Qwen3.5 9B}, yet it shows the strongest interaction between them; the interaction is similar for \mbox{EuroLLM 22B} but negligible for both Gemma-4 models. Across models, ordering matters more when the demonstrations are class-balanced, whereas its effect is negligible under the \emph{proportional} ratio---except for \mbox{Qwen3.5 9B}.

\paragraph{Static Demonstrations.}
The sets of demonstrations selected with the random search are reported in Appendix~\ref{app:static-demos}. Under the static configuration, \mbox{Gemma-4 26B} reaches an $F1_{\text{macro}}$ of $0.723$ for \emph{Title} conditioning and an $F1_{\text{macro}}$ of $0.754$ for \emph{Implicit} conditioning, beating the results of best \emph{Title} ($+ 0.028$) and \emph{Implicit} ($+ 0.021$) conditioning configuration of the main experiments.

\section{Discussion}\label{sec:discussion}

Although RetICL is our focus, the retrieval mechanisms contribute remarkably little to model performance: embedding mode and retrieval strategy move $F1_{\text{macro}}$ far less than the conditioning mode, the zero-~/~few-shot distinction, or the number, class ratio, and ordering of demonstrations, and the elaborate dynamic strategies barely beat \emph{random} retrieval. This nuances the endorsement of similarity-based selection by \citet{margatina-etal-2023-active} and aligns with \citet{liu2024poliprompt}: for criminally relevant hate speech, the gains demonstrations bring primarily come from few-shot prompting itself, not from how examples are selected. Also consistent with \citet{liu2024poliprompt} is that an optimised, static set of demonstrations improves performance over all dynamic approaches within the same conditioning mode (at least for the tested configurations on \mbox{Gemma-4 26B}), suggesting that good demonstration selection for this task depends on metrics other than those used for dynamic selection. The fragility of fine-grained demonstration tuning is underlined by the shared task results: the ratio~/~ordering configuration that won in cross-validation fell behind the default on the held-out test set (Appendix~\ref{app:shared-task}).

Model choice matters more than prompting: its mean impact on $F1_{\text{macro}}$ ($0.309$) exceeds that of any single prompting axis (Fig.~\ref{fig:impact_prompting}). Scale, however, is not decisive---the small \mbox{Gemma-4 E4B} ($0.659$) beats the larger \mbox{EuroLLM 22B} ($0.474$), and even GPT-5.5 collapses to $0.440$ under \emph{Title} zero-shot prompting before reaching a result of $0.716$ with \emph{Explicit} conditioning. Weaker models are also often more configuration-sensitive (\mbox{EuroLLM 22B} spans ${\sim}0.5$ $F1_{\text{macro}}$ across all its runs, including ablations), so a poor model should not be dismissed before its prompting is tuned; capability tracks training and conditioning rather than size \citep{miller_dynamic_2025}. 

Contrary to \citet{ludwig_conditioning_2025}, who found abstract conditioning superior for §\,130 StGB, more concrete, decomposed legal knowledge helps here: \emph{Explicit} conditioning is best on average and best for three of four models, while \emph{Title} is usually worst. The effect is neither strictly monotonic nor model-independent, supporting subtask decomposition \citep{alkhamissi2022token} while showing that the right form of legal conditioning depends on the model. Decomposition also aids explainability, a further argument for this configuration.

The naive TF-IDF baseline ($F1_{\text{macro}}=0.644$) is strikingly competitive, outperforming most smaller-model runs. This may reflect homogeneous data, where lexical overlap already proxies the label well---a shortcut unlikely to survive distribution shift or concept drift, where LLMs' semantic abilities should matter more. 
Reliability is a further concern: Recall on the positive class exceeds Precision, so the systems over-predict criminal relevance yet still miss between $26\%$ and $57\%$ of criminally relevant posts, positioning these models as triage tools for human review rather than for unsupervised moderation. Under this framing, the models clearly beat the baseline, which misses $64$\% of criminally relevant posts.
Over-sensitivity of LLMs may be an unintended side-effect of alignment: \citet{selvaganapathy-nasim-2026-confident} find that minimally aligned models outperform aligned ones in hate speech classification, echoing a broader unintended trade-off between safety and truthfulness reported by \citet{mahmoud-etal-2026-unintended}. 

Finally, our QLoRA fine-tune of the small \mbox{Gemma-4 E4B} ($F1_{\text{macro}}=0.741$) exceeds the best dynamic prompting configuration ($0.733$, \mbox{Gemma-4 26B}), a lead that holds on the held-out test set ($0.74$ vs.\ $0.72$): a cheaply fine-tuned 4B model can match a carefully prompted 26B one, so where a modest labelled set exists, parameter-efficient fine-tuning is a strong alternative to ICL. However, it is less suited as a triage tool, as it misses more than fifty percent of defamatory offences, albeit being more precise when predicting the positive class.

\section{Conclusion}\label{sec:conclusion}
We tested prompting strategies for detecting defamatory offences under \mbox{§§\,185--187 StGB}, varying both the legal knowledge supplied to the model and the way few-shot demonstrations are retrieved. Contrary to \citet{ludwig_conditioning_2025}, more concrete legal knowledge helps; and while few-shot prompting clearly outperforms zero-shot, more elaborate retrieval offers only a marginal advantage over randomly chosen demonstrations. Optimised static demonstrations can outperform dynamic retrieval, at least in the settings we tested. Model choice has more impact on task performance than any prompting configuration axis alone, and the task remains hard: most of the open-weights models we tested do not consistently match a naive nearest-neighbour baseline in terms of $F1_{\text{macro}}$. However, models outperform this baseline consistently in terms of Recall, which we argue is the more relevant metric for a triage setting.
The interaction of demonstration class ratio and ordering, the way such positional and class biases are represented within a model, and their stability across data splits remain promising directions for future work.

\section*{Limitations}
Our findings are confined to a single language, dataset, and annotation scheme, all drawn from the same distribution; the homogeneity that makes the nearest-neighbour baseline strong also limits external validity. Assessing how concept drift and more heterogeneous data affect the baseline, the RetICL strategies, and the fine-tuning approach is a natural next step---we conjecture that nearest-neighbour, SFT and static demonstration approaches, which lean on specific training instances, may degrade more than concept-driven LLM prompting under such shift, though this remains to be tested. We also did not exhaust the entire configuration space: the most favourable demonstration settings were not all combined. Reasoning-based legal prompting (\emph{legal syllogism}~/~chain-of-thought; \citealp{kepu-2025-syler, jiang-2023-legal, deng-etal-2024-enabling}) remains an underexplored extension for explicitly modelling the statutory decision tree, as we only use models with their respective \emph{thinking modes} disabled. Finally, §§\,185--187 are usually prosecuted only on application by the victim (§\,194 StGB); proactive automatic detection is therefore of more limited practical relevance for these offences than for \emph{ex officio} offences such as §\,130 StGB (incitement to hatred), where the public prosecutor initiates proceedings on its own motion \citep{beulke2025strafprozessrecht}. 
%Finally, §§\,185--187 are prosecuted only on application by the victim (§\,194 StGB) and only exceptionally \emph{ex officio}\footnote{Under the principle of \emph{ex officio} prosecution (\emph{Offizialprinzip}, §\,152(1) StPO), the public prosecutor initiates proceedings on its own motion, independently of a victim's complaint; complaint offences (\emph{Antragsdelikte}) are the exception \citep{beulke2025strafprozessrecht}.}; proactive automatic detection is therefore of more limited practical relevance for these offences than for \emph{ex officio} offences such as §\,130 StGB (incitement to hatred). 

%\section*{Acknowledgments}

\clearpage

\bibliography{custom}

\clearpage

\appendix

\section*{Appendix}

\section{Annotation Scheme}\label{app:annotation-scheme}

The dataset of the shared task is labelled according to the decision scheme of \citet{zufall-etal-2019-legal} for \mbox{§§\,185--187 StGB} (\autoref{subsec:annotation-scheme}). The three offences differ mainly in the nature of the statement and its addressee \citep{bundestag2022}:

\begin{itemize}[noitemsep]
    \item \textbf{§\,185} \emph{Insult} (\emph{Beleidigung}): expressions of contempt toward a person, typically value judgments that can be proven neither true nor false.
    \item \textbf{§\,186} \emph{Defamation} (\emph{Üble Nachrede}): asserting a disparaging \emph{fact} about someone to third parties when its truth cannot be established.
    \item \textbf{§\,187} \emph{Intentional Defamation} (\emph{Verleumdung}): the aggravated case of knowingly spreading a false fact, carrying the highest penalty.
\end{itemize}

The scheme of \citet{zufall-etal-2019-legal} operationalises punishability as six binary (yes~/~no) decisions taken solely from the text of a post: a three-step spine (steps~1--3) and, for value judgments, a three-way balancing of competing rights (steps~4--6).

\begin{enumerate}[noitemsep]
    \item \textbf{Defamatory object.} The post must target a valid holder of the right to honour: a \emph{living individual} \emph{or} a \emph{specific group} distinguishable from the general public (including collective entities such as governments or companies). If neither is addressed, the statement is \emph{not punishable}.
    \item \textbf{Disparaging statement.} The statement must express contempt or allege shortcomings that could lower the victim's social standing. If it is not disparaging, it is \emph{not punishable}.
    \item \textbf{Value judgment vs.\ factual claim.} A disparaging statement is either a factual claim (provably true or false) or a value judgment (personal opinion). Establishing the punishability of a \emph{factual claim} would require the court to take evidence on its truth---information beyond the text---so the assessment halts here. \emph{Value judgments} proceed to the balancing step.
\end{enumerate}

\noindent\textbf{Balancing of rights (steps~4--6; value judgments only).} A value judgment may be protected under freedom of expression and thus be safeguarded as \textit{legitimate interest} (§\,193 StGB). The scheme encodes the established case-law for weighing this freedom against the right to honour as three further binary decisions:

\begin{enumerate}[noitemsep,resume]
    \item \emph{Abusive insult} (\emph{Formalbeleidigung}): a taboo-breaking statement intended only to defame, excluded from free-speech protection $\rightarrow$ \emph{punishable}.
    \item \emph{Topic of public interest}: a contribution to public discourse, carrying a presumption in favour of free speech $\rightarrow$ \emph{usually not punishable}.
    \item \emph{Abusive criticism} (\emph{Schmähkritik}): a statement going beyond plausible criticism primarily to offend the victim $\rightarrow$ \emph{usually punishable}.
\end{enumerate}

Where these signals conflict or coincide, the outcome depends on a free judicial balancing of the individual circumstances, which \citet{zufall-etal-2019-legal} deliberately leave unimplemented. 
The task, however, requires strictly binary True~/~False labels and does not allow for an ``undecided'' class. We therefore resolved undecided steps to ``not punishable'', following the principle \textit{in dubio pro reo} (``when in doubt, for the accused''). An alternative would be to resolve undecided steps to ``punishable'' instead---a defensible choice if the scheme is framed as a triage tool, as we argue for in the main part of this paper, where the goal is to flag potentially problematic content for human review rather than to render a final legal judgment.

Relatedly, while \citet{zufall-etal-2019-legal} operationalise the \textit{punishability} of a post, we frame the task more broadly in this paper as classifying posts with respect to their \textit{criminal relevance} under \mbox{§§\,185--187 StGB}. Since we are not legal experts and frame the task as a triage tool for law enforcement or content moderation rather than a substitute for judicial assessment, we deliberately leave any definitive judgement of punishability or prosecutability to human experts downstream.

\citet{zufall-etal-2019-legal} report that suitably instructed laypeople can apply the scheme with reasonable reliability relative to an expert reference, though they tend to interpret \emph{specific group} and \emph{disparaging statement} more leniently.

\section{Conditioning}
\label{app:prompts}

\definecolor{pineblue}{RGB}{83, 128, 131}

\subsection{Content}

\begin{tcolorbox}[breakable,colback=pineblue!10!white, colframe=pineblue,boxsep=2pt, left=3pt, right=3pt, top=2pt, bottom=2pt, arc=3pt,title=System Prompt]
You are a legal expert for defamatory offences according to the German Criminal Code (\mbox{§§\,185--187 StGB}). Help the user decide whether given posts fall within the scope of these sections.\end{tcolorbox}

\paragraph{Task}

All task descriptions are followed by the prompt to ``Answer only with 'True' (criminally relevant according to \mbox{§§\,185--187 StGB}) or 'False' (not criminally relevant according to \mbox{§§\,185--187 StGB}).'', or ``Answer only with 'True' or 'False'.'' for the explicit conditioning mode.

%\paragraph{Title}
\begin{tcolorbox}[breakable, colback=pineblue!10!white, colframe=pineblue,boxsep=2pt, left=3pt, right=3pt, top=2pt, bottom=2pt, arc=3pt,title=Title]
Is the following post criminally relevant with respect to \mbox{§§\,185--187 StGB} (Defamatory Offence)?
\end{tcolorbox}

\begin{tcolorbox}[breakable, colback=pineblue!10!white, colframe=pineblue,boxsep=2pt, left=3pt, right=3pt, top=2pt, bottom=2pt, arc=3pt,title=Description]
Is the following post criminally relevant with respect to \mbox{§§\,185--187 StGB} (Defamatory Offence)? A statement is criminally relevant under these titles only if either a living individual or a specific group is an object of the respective statement, if the statement is disparaging, the statement is a value judgement and not a factual claim. If these hold and the statement constitutes an abusive insult, it is criminally relevant. If it is not an abusive insult, it is only criminally relevant if it contains abusive criticism and if the topic is not of public interest.\end{tcolorbox}

\begin{tcolorbox}[breakable, colback=pineblue!10!white, colframe=pineblue,boxsep=2pt, left=3pt, right=3pt, top=2pt, bottom=2pt, arc=3pt,title=Implicit Steps]
Step 1: A defamatory object can be a living individual, a group of persons that is distinguishable from the general public such that every member of that group could feel their honour is infringed or collective entities such as governments or press companies with a recognised social role and who act with a collective, single will.\\
Is there a defamatory object in this statement? \\
No → STOP (NOT criminally relevant according to \mbox{§§\,185--187 StGB})\\
Yes → Step 2\\

Step 2: A disparaging statement is a statement which interferes with the potential victim's right to honour. It is already fulfilled by expressing contempt or disrespect through the allegation of shortcomings that could reduce the victim’s social standing.\\
Is there a disparaging statement directed towards the defamatory object?\\
No → STOP (NOT criminally relevant according to \mbox{§§\,185--187 StGB})\\
Yes → Step 3\\

Step 3: A factual claim is a statement that can be proven to be true or untrue in front of a court. A value judgment constitutes an expression of personal opinions.\\
Is it primarily a value judgement and NOT primarily a factual claim?\\
No → STOP (NOT criminally relevant according to \mbox{§§\,185--187 StGB})\\
Yes → Step 4\\

Step 4: An abusive insult is a statement that constitutes breaking a taboo by itself and intends only the defamation of the victim without any substantiated contribution.\\
Is the statement an abusive insult?\\
Yes → STOP (criminally RELEVANT according to \mbox{§§\,185--187 StGB})\\
No → Step 5\\

Step 5: A statement is of public interest if it contains a contribution to the public discourse with respect to a particular relevant topic of public interest.\\
Is the statement of public interest?\\
Yes → STOP (NOT criminally relevant according to \mbox{§§\,185--187 StGB})\\
No → Step 6\\

Step 6: A statement is considered abusive criticism if it goes beyond plausible criticism by primarily intending to abusively offend the victim, hereby neglecting a substantiated contribution.\\
Is the statement abusive criticism?\\
Yes → STOP (criminally RELEVANT according to \mbox{§§\,185--187 StGB})\\
No → STOP (NOT criminally relevant according to \mbox{§§\,185--187 StGB})\\

According to this decision scheme, decide whether the following text is criminally relevant according to \mbox{§§\,185--187 StGB} (Defamatory Offence).\end{tcolorbox}

\begin{tcolorbox}[breakable, colback=pineblue!10!white, colframe=pineblue,boxsep=2pt, left=3pt, right=3pt, top=2pt, bottom=2pt, arc=3pt,title=Explicit]

\begin{tcolorbox}[breakable, colback=pineblue!10!white, colframe=pineblue,boxsep=2pt, left=3pt, right=3pt, top=2pt, bottom=2pt, arc=3pt,title=Step 1]
A defamatory object can be a living individual, a group of persons that is distinguishable from the general public such that every member of that group could feel their honour is infringed or collective entities such as governments or press companies with a recognised social role and who act with a collective, single will.\\
Is there a defamatory object in this statement?\end{tcolorbox}

\begin{tcolorbox}[breakable, colback=pineblue!10!white, colframe=pineblue,boxsep=2pt, left=3pt, right=3pt, top=2pt, bottom=2pt, arc=3pt,title=Step 2]
A disparaging statement is a statement which interferes with the potential victim's right to honour. It is already fulfilled by expressing contempt or disrespect through the allegation of shortcomings that could reduce the victim’s social standing.\\
Is there a disparaging statement directed towards the defamatory object? \end{tcolorbox}

\begin{tcolorbox}[breakable, colback=pineblue!10!white, colframe=pineblue,boxsep=2pt, left=3pt, right=3pt, top=2pt, bottom=2pt, arc=3pt,title=Step 3]
A factual claim is a statement that can be proven to be true or untrue in front of a court. A value judgment constitutes an expression of personal opinions.\\
Is the statement primarily a value judgement? \end{tcolorbox}

\begin{tcolorbox}[breakable, colback=pineblue!10!white, colframe=pineblue,boxsep=2pt, left=3pt, right=3pt, top=2pt, bottom=2pt, arc=3pt,title=Step 4]
An abusive insult is a statement that constitutes breaking a taboo by itself and intends only the defamation of the victim without any substantiated contribution.\\
Is the statement an abusive insult? \end{tcolorbox}

\begin{tcolorbox}[breakable, colback=pineblue!10!white, colframe=pineblue,boxsep=2pt, left=3pt, right=3pt, top=2pt, bottom=2pt, arc=3pt,title=Step 5]
A statement is of public interest if it contains a contribution to the public discourse with respect to a particular relevant topic of public interest.\\
Is the statement of public interest?\end{tcolorbox}

\begin{tcolorbox}[breakable, colback=pineblue!10!white, colframe=pineblue,boxsep=2pt, left=3pt, right=3pt, top=2pt, bottom=2pt, arc=3pt,title=Step 6]
A statement is considered abusive criticism if it goes beyond plausible criticism by primarily intending to abusively offend the victim, hereby neglecting a substantiated contribution.\\
Is the statement abusive criticism?\end{tcolorbox}
\end{tcolorbox}

\subsection{Chat Template}

All experiments were conducted with \emph{user}-\emph{assistant} turns using the models' tokenisers' chat templates. The prompts are structured as follows:

\begin{tcolorbox}[breakable, colback=pineblue!10!white, colframe=pineblue,boxsep=2pt, left=3pt, right=3pt, top=2pt, bottom=2pt, arc=3pt]
\texttt{[\{role:system,content:\textbf{system prompt}\},\\
\{role:user,content:\textbf{task}\},\\
\{role:user,content:\textbf{demonstration\_1}\},\\
\{role:assistant,content:\textbf{class\_1}\},\\
\{role:user,content:\textbf{demonstration\_2}\},\\
\{role:assistant,content:\textbf{class\_2}\},\\
...\\
\{role:user,content:\textbf{demonstration\_k}\},\\
\{role:assistant,content:\textbf{class\_k}\},\\
\{role:user,content:post\}]}
\end{tcolorbox}

\subsection{Static Demonstrations}\label{app:static-demos}

\begin{tcolorbox}[breakable, colback=red!10!white, colframe=red,boxsep=2pt, left=3pt, right=3pt, top=2pt, bottom=2pt, arc=3pt,title=Content Warning]
This section contains examples of hate speech, including instances that are highly offensive and, in some cases, criminally relevant under German law (\mbox{§\,130}, \mbox{§§\,185--187 StGB}, inter alia). These examples are included solely for scientific and illustrative purposes, to enable transparent evaluation and reproducibility of our methods. They do not reflect the views of the authors. Reader discretion is advised.
\end{tcolorbox}

The sets of demonstrations optimised using a random search (see Appendix~\ref{app:random-search}) are reported here.

\begin{tcolorbox}[breakable, colback=pineblue!10!white, colframe=pineblue,boxsep=2pt, left=3pt, right=3pt, top=2pt, bottom=2pt, arc=3pt,title=Title Conditioning]
"... komisches Interview... Hallo Baden-Württemberger: was hat euch dazu bewogen, diese Spastiker zu wählen ???...."\\ \textbf{True}\\
"Die ehrlose faschistische Ratte habe ich nie gekannt, sie ist ausgelöscht aus meinem Leben."\\ \textbf{True}\\
"Im Sommer werden solche Journalisten gehängt ;)"\\ \textbf{False}\\
"Linksfaschos schlagen auf Grenzer ein damit Flüchtlinge aus dem Herkunftsland der Linksfaschos ausreisen können? Ja wie bekloppt ist das denn? Das ist doch völlig absurd!"\\ \textbf{False}\\
"Ob Journalisten oder Politiker, es sind doch alle nur noch scheinheilige Geister. Wenn sich diese sogenannte feine Gesellschaft anmaßt uns als Nazi, Mischpoke und Pack zu betiteln, dann werden wir uns doch erlauben dürfen dieses miese etwas in ähnlicher Weise zu behandeln."\\ \textbf{False}\\
"SOLLEN FROH SEIN ÜBERHAUPT HIER SEIN ZU DÜRFEN, DRECKSGELUMPE"\\ \textbf{False}\\
"Wer keine Ahnung von der Realität hat sollte die Fresse halten, kleines Antideutsches [name] Würstchen :D [url]"\\ \textbf{True}\\
"[name] du roter vollpfosten"\\ \textbf{True}
\end{tcolorbox}

\begin{tcolorbox}[breakable, colback=pineblue!10!white, colframe=pineblue,boxsep=2pt, left=3pt, right=3pt, top=2pt, bottom=2pt, arc=3pt,title=Implicit Conditioning]
"@user, du bist geistig behindert und ein saudummer AFD-Russentroll !!!"\\ \textbf{True}\\
"Grammatik bei Yoda gelernt du hast ! Was für ein unterbelichtetes Höhlenkind. And now: GTFO !"\\ \textbf{True}\\
"Krankes Pack!"\\ \textbf{False}\\
"[name] Du kleiner dummer Junge,hast keine Ahnung vom Leben.Ich möchte dir so gerne eine neue saubere Windel überziehen.Die hast du dringend nötig!"\\ \textbf{True}\\
"Solange [name] keine nachprüfbare Dissertation vorlegen kann, ist er für mich der größte Scharlatan und Volksschädling aller Zeiten. Es ist an der Zeit für ihn Gift oder Kugel zu wählen. Der angerichtete Schaden für Wiedergutmachung ist zu groß."\\ \textbf{True}\\
"War doch eh wieder ein Asylbewerber...vielleicht gings ja wieder um die Haushaltsführung??? Kann [name] und Konsorten wieder Hände schütteln gehen und alles auf die bösen Nazis schieben. Die Armen brauchen nämlich Trost und Schutz. Am besten nimmt die [name] und [name] und wie sie alle heißen gleich noch ein paar Nutten mit, natürlich zum Nulltarif... Die nordafrikanischen Testosteronbolzenschäumen sonst über und das wollen wir ja nicht...nicht das wir noch mit dem europäischen Gerichtshof für Menschenrechte zu tun bekommen..."\\ \textbf{False}\\
"also ich würde fahren und wenn das blut erstmal 5meter weit spritzt knochen und körper fliegen  machen die anderen platz haben auf der bab nichts aber auch garnichts zu suchen ( eingriff in straßenverkehr )"\\ \textbf{False}\\
"Über die Ossis damals gemault! Die schwarzen willkommen heißen! Ihr seid ja Bazis"\\ \textbf{False}\\
\end{tcolorbox}

\section{Detailed Experiment Settings}\label{app:detailed-settings}

Except for the API-based inference calls, all models are loaded from the HuggingFace Hub and run with the Transformers library \citep{wolf-etal-2020-transformers}. Inference calls are executed in batches for quicker computation. \emph{Thinking} mode is explicitly turned off for the models and \emph{max\_tokens} is set to $10$, because we expect only ``True'' or ``False'' labels in models' generated answers.

For the ablation using GPT-5.5, \emph{max\_tokens} has to be changed because minimum \emph{max\_tokens} when using the API is $16$, which we set accordingly.

Generated answers are parsed directly as True~/~False labels, not allowing for any additional text. This works very well across all models; fewer than $1\%$ of answers cannot be parsed (see Appendix~\ref{app:abstentions}).

\subsection{Exploration Setup}\label{app:exploration}

When exploring demonstration sizes \mbox{$k\in\{4,8,16,32\}$}, each model~/~$k$ combination is explored twice: once with a \emph{Title}, \emph{dynamic}, \emph{dense}, \emph{similarity} prompting configuration and once with \emph{Description}, \emph{dynamic}, \emph{random} configuration.

\subsection{Knowledge Bases}\label{app:knowledge-bases}

\begin{figure}[ht]
    \centering
    \includegraphics[width=\linewidth]{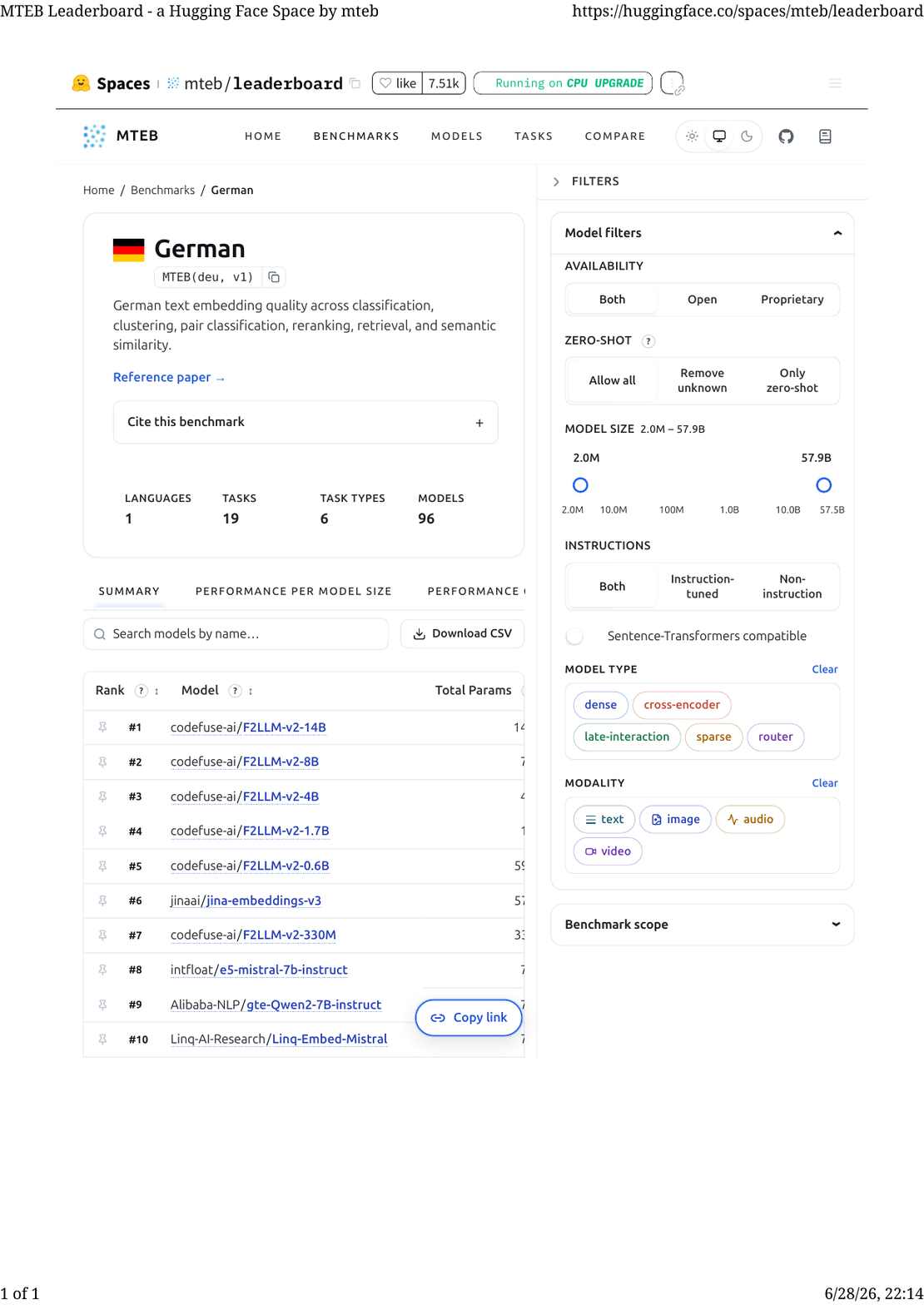}
    \caption{Screenshot of the MTEB Leaderboard for embedding models on German tasks, dated 28 June 2026.}
    \label{fig:mteb}
\end{figure}

For the \emph{dense} embeddings, the \texttt{codefuse-ai/F2LLM-v2-1.7B} \citep{f2llm-v2} embedding model is used, which ranked among the highest on the MTEB leaderboard for German tasks (Fig.~\ref{fig:mteb}) during the first half of 2026. Only larger models of the same family surpass this model, but for efficiency reasons we opt for the 1.7-billion model.

When using both \emph{similarity}- and \emph{MMR}-based retrieval, these embeddings are used to build a LangChain vector store\footnote{\url{https://reference.langchain.com/python/langchain-core/vectorstores/in_memory/InMemoryVectorStore}}, which is queried at inference time for selecting the best training samples based on the respective retrieval strategy. LangChain's vector store enables filtering when querying, making it easy to retrieve samples based on their class.

For \emph{diversity}-based retrieval, the dense embeddings are clustered using sklearn's \texttt{K-Means} algorithm, where $K=\frac{\#demonstrations}{2}$. Clusters are built separately for positive and negative classes. At inference time, a training sample is drawn at random from each cluster, resulting in a diverse set of demonstrations, balanced by class.

The \emph{sparsely} embedded knowledge base is built with a \emph{BM25Retriever} module\footnote{\url{https://reference.langchain.com/python/langchain-community/retrievers/bm25/BM25Retriever}}, a LangChain community build. Because this retriever does not allow for filtered similarity search, a retriever is built for each class separately. At inference time, the most similar training samples are retrieved for both classes.

\subsection{Explicit Conditioning}\label{app:explicit-cond}

The \emph{Explicit} mode deconstructs the task into six subtasks. These are prompted sequentially. After each step, the model's answer is parsed into a \texttt{True} or \texttt{False} label. If the decision at step $i$ results in a definite final label (e.g. if there is no defamatory object present (step $1$), resulting in a final \texttt{False} label), the decision scheme is halted and the final label is set.

The additionally labelled dataset is created in the same fashion---resulting in single-step annotations only for the subset of steps until a final decision is reached. When using the \emph{Explicit} mode in a few-shot setting, this dataset is used for the demonstration pool. Labels get rarer for later steps, and there is a strong class imbalance for some of the steps (in total, there are only $10$ positive instances for \emph{abusive insult}). Especially for the clustering method of the \emph{diversity}-based retrieval, this can mean that not enough demonstrations for one label can be retrieved at inference time. In the rare cases that this happens, demonstration \emph{ratio} takes precedence over demonstration \emph{size}: the demonstrations will be chosen so that the resulting set of few-shot demonstrations is \emph{balanced}, even if this means that fewer than $8$ demonstrations are used.

\subsection{Proprietary Model}\label{app:proprietary}

The GPT-5.5 model is used with the dated snapshot version \texttt{gpt-5.5-2026-04-23}.

\subsection{Fine-Tuning}\label{app:fine-tuning}

\mbox{Gemma-4 E4B} is loaded with 4-bit NF4 quantisation and double quantisation, and LoRA adapters are trained on the attention and feed-forward projection layers. Training uses a maximum sequence length of 1024 tokens, three epochs, batch size 1 with 32 gradient accumulation steps, a learning rate of $2 \times 10^{-5}$, and gradient checkpointing. Prompt tokens are masked from the loss so that optimisation targets only the label completion. Generation uses \texttt{thinking\_mode: False} with a maximum budget of 10 tokens per sample.

\subsection{Static Demonstrations}\label{app:random-search}

To select the demonstration set for the static few-shot condition, we perform a random search over candidate sets drawn from the training data. We sample $30$ class-balanced candidate sets of $k=8$ demonstrations each ($k/2$ per class) and evaluate every set by using it as few-shot context for classifying a fixed evaluation subset of $300$ posts, sampled from the training data stratified by class label and kept disjoint from the demonstration pool. Each candidate set is scored by $F1_{\text{macro}}$ over the model's predictions; the best-scoring set is retained. The search is conducted separately for the two prompt templates \emph{Title} and \emph{Implicit}, so the selected demonstrations are optimised for the exact prompt they are later used with. We only optimise sets of demonstrations for the \mbox{Gemma-4 26B} model.

\section{Detailed Results}\label{app:results}

Full results of all runs of the main experiments are reported in Table~\ref{tab:full}.

\subsection{Exploration Results}\label{app:exploration-results}

Table~\ref{tab:exploration-demo-size} reports the mean $F1_{\text{macro}}$ scores per model, for all explored demonstration sizes \mbox{$k \in \{4,8,16,32\}$}, along with the respective gain or loss in $F1_{\text{macro}}$ score after doubling the demonstration size.

\begin{table}[!ht]
\centering
\resizebox{.48\textwidth}{!}{
\begin{tabular}{lrrrr}
\toprule
Model & $k$=4 & $k$=8 & $k$=16 & $k$=32 \\
\midrule
Gemma-4 26B & 0.660 & 0.689 (\textbf{+0.029}) & 0.682 (-0.007) & 0.686 (+0.005) \\
Gemma-4 E4B & 0.547 & 0.596 (\textbf{+0.049}) & 0.631 (+0.035) & 0.673 (+0.042) \\
Qwen3.5 9B & 0.549 & 0.560 (+0.011) & 0.584 (+0.024) & 0.617 (\textbf{+0.033}) \\
EuroLLM 22B & 0.373 & 0.399 (+0.026) & 0.443 (\textbf{+0.044}) & 0.475 (+0.031) \\
\midrule
Mean & 0.532 & 0.561 (\textbf{+0.029}) & 0.585 (+0.024) & 0.613 (+0.028) \\
\bottomrule
\end{tabular}
}
\caption{Mean $F1_{\text{macro}}$ per model and $k$. Values in parentheses show the gain~/~loss vs. the previous $k$. Largest gains are reported in \textbf{bold}.}\label{tab:exploration-demo-size}
\end{table}

\subsection{RetICL Configurations}\label{app:retrieval}

In the paper, we show that the \emph{embedding} and \emph{retrieval} modes have very little impact on model performance in comparison to \emph{conditioning} mode and the difference between zero- and few-shot. 

The best configuration per model (Table~\ref{tab:e-best-per-model}) is \emph{similarity}-based for all models, based on \emph{dense} embeddings for both \mbox{Gemma-4 26B} and \mbox{Qwen3.5 9B} and based on \emph{sparse} embeddings for \mbox{Gemma-4 E4B} and \mbox{EuroLLM 22B}. However, on average (Table~\ref{tab:a3-model-retrieval}), a knowledge base made up of \emph{dense} embeddings performs best for all models. While \emph{similarity}-based retrieval is also best on average for both Gemma-4 models and \mbox{EuroLLM 22B}, \emph{diversity}-based retrieval performs slightly better on average for \mbox{Qwen3.5 9B}.

While \emph{dense} \emph{similarity}-based RetICL performs well on average, these settings should be explored for each \emph{conditioning} mode so that the best overall configuration can be found.

\begin{table}[ht]
\centering
\resizebox{.48\textwidth}{!}{
\begin{tabular}{lrrrrrrr}
\toprule
Model & D-Sim & D-MMR & D-Div & F-Sim & S-Sim & Rand & $\Delta$ \\
\midrule
Gemma-4 26B & \textbf{0.709} & 0.709 & 0.701 & 0.705 & \textit{0.693} & 0.696 & 0.016 \\
Gemma-4 E4B & \textbf{0.607} & \textit{0.584} & 0.606 & 0.598 & 0.594 & 0.600 & 0.023 \\
Qwen3.5 9B & 0.579 & \textit{0.576} & \textbf{0.596} & 0.579 & 0.583 & 0.595 & 0.019 \\
EuroLLM 22B & \textbf{0.438} & 0.401 & \textit{0.384} & 0.427 & 0.430 & 0.388 & 0.054 \\
\bottomrule
\end{tabular}
}
\caption{$F1_{\text{macro}}$ by model and retrieval configuration (k=8, mean over all prompt modes). \textbf{Bold}: best retrieval config per model; \textit{italic}: worst. $\Delta$ = best$-$worst.} \label{tab:a3-model-retrieval}
\end{table}

\subsection{Abstentions}\label{app:abstentions}

Absolute and relative abstention rates are reported in Table~\ref{tab:abstentions}. Abstention is very low overall, with median $0$--$1$ abstentions per run, which is less than $1\%$ of posts.
For \mbox{Gemma-4 26B}, $648$ of the $657$ abstentions occur in a single configuration: under \emph{Explicit} zero-shot prompting, the model asks for further information to classify the first step (\emph{defamatory object}) instead of replying \texttt{True} or \texttt{False}. Without context, many posts are ambiguous at this step, since personal pronouns---especially in the plural---cannot conclusively establish the existence of a defamatory object. When demonstrations are supplied as user--assistant turns, the model is far more likely to commit to an answer despite the ambiguity.

\begin{table}[ht]
\centering
\small
\begin{tabular}{lrrrr}
\toprule
Model & Total & Mean & Median & Rate \\
\midrule
Gemma-4 26B & 657 & 23.5 & 0.0 & 0.7\%\\
Gemma-4 E4B & 14 & 0.5 & 0.0 & 0.02\%\\
Qwen3.5 9B & 25 & 0.9 & 1.0 & 0.03\%\\
EuroLLM 22B & 25 & 0.9 & 1.0 & 0.03\%\\
\bottomrule
\end{tabular}
\caption{Abstentions per model across all configurations (total, mean and median per config and total rate).}\label{tab:abstentions}
\end{table}

\subsection{Class Ratio and Ordering}\label{app:results-ratio-ordering}

% \begin{table}[ht]
% \centering
% \resizebox{.48\textwidth}{!}{
% \setlength{\tabcolsep}{4pt}
% \begin{tabular}{lrrrrrr}
% \toprule
% Model & \multicolumn{3}{c}{Balanced} & \multicolumn{3}{c}{Proportional} \\
% \cmidrule(lr){2-4} \cmidrule(lr){5-7}
%  & Random & True first & True last & Random & True first & True last \\
% \midrule
% Gemma-4 26B & 0.723 & 0.740 & \textit{0.706} & 0.727 & \textbf{0.741} & 0.728 \\
% Gemma-4 E4B & 0.579 & 0.592 & \textit{0.540} & 0.677 & \textbf{0.690} & 0.665 \\
% Qwen3.5 9B & 0.604 & \textbf{0.704} & \textit{0.581} & 0.663 & 0.608 & 0.653 \\
% EuroLLM 22B & 0.445 & 0.546 & \textit{0.361} & \textbf{0.663} & 0.656 & 0.653 \\
% \bottomrule
% \end{tabular}
% }
% \caption{$F1_{\text{macro}}$ by demonstration class ratio and ordering (balance $\times$ order ablation). \textbf{Bold}: best combination per model; \textit{italic}: worst.}\label{tab:ablation-balance-order}
% \end{table}

\begin{table*}[t]
\centering
\resizebox{.75\textwidth}{!}{
\setlength{\tabcolsep}{4pt}
\begin{tabular}{llrrrrrrrrr}
\toprule
&  & \multicolumn{3}{c}{Random} & \multicolumn{3}{c}{True first} & \multicolumn{3}{c}{True last} \\
\cmidrule(lr){3-5} \cmidrule(lr){6-8} \cmidrule(lr){9-11}
 &  & F1 & P$_\mathsf{T}$ & R$_\mathsf{T}$ & F1 & P$_\mathsf{T}$ & R$_\mathsf{T}$ & F1 & P$_\mathsf{T}$ & R$_\mathsf{T}$ \\
\midrule
Gemma-4 26B & Balanced & 0.723 & 0.429 & 0.733 & 0.740 & 0.482 & 0.663 & 0.706 & 0.396 & \textbf{0.764} \\
 & Proportional & 0.727 & 0.471 & 0.617 & \textbf{0.741} & \textbf{0.523} & 0.586 & 0.728 & 0.465 & 0.641 \\
\addlinespace
Gemma-4 E4B & Balanced & 0.579 & 0.252 & 0.802 & 0.592 & 0.262 & \textbf{0.815} & 0.540 & 0.222 & 0.788 \\
 & Proportional & 0.677 & 0.393 & 0.528 & \textbf{0.690} & \textbf{0.415} & 0.547 & 0.665 & 0.382 & 0.484 \\
\addlinespace
Qwen3.5 9B & Balanced & 0.604 & 0.270 & 0.740 & \textbf{0.704} & 0.466 & 0.511 & 0.581 & 0.252 & \textbf{0.793} \\
 & Proportional & 0.663 & 0.547 & 0.306 & 0.608 & \textbf{0.670} & 0.176 & 0.653 & 0.362 & 0.465 \\
\addlinespace
EuroLLM 22B & Balanced & 0.445 & 0.181 & 0.870 & 0.546 & 0.225 & 0.773 & 0.361 & 0.160 & \textbf{0.918} \\
 & Proportional & \textbf{0.663} & 0.397 & 0.436 & 0.656 & \textbf{0.619} & 0.270 & 0.653 & 0.337 & 0.583 \\
\bottomrule
\end{tabular}}
\caption{$F1_{\text{macro}}$ and true-class precision (P$_\mathsf{T}$) and recall (R$_\mathsf{T}$) for every demonstration ratio $\times$ order combination (balance $\times$ order ablation). \textbf{Bold}: best value per model across all combinations.}
\label{tab:ablation-pr-grid}
\end{table*}

\begin{table*}[t]
\centering
\resizebox{.98\textwidth}{!}{
\setlength{\tabcolsep}{4pt}
\begin{tabular}{l|rrrrrr|rrrrrrrrr}
\toprule
& \multicolumn{3}{c}{Balanced} & \multicolumn{3}{c|}{Proportional} & \multicolumn{3}{c}{Random} & \multicolumn{3}{c}{True first} & \multicolumn{3}{c}{True last} \\
\cmidrule(lr){2-4} \cmidrule(lr){5-7} \cmidrule(lr){8-10} \cmidrule(lr){11-13} \cmidrule(lr){14-16}
 & F1 & P$_\mathsf{T}$ & R$_\mathsf{T}$ & F1 & P$_\mathsf{T}$ & R$_\mathsf{T}$ & F1 & P$_\mathsf{T}$ & R$_\mathsf{T}$ & F1 & P$_\mathsf{T}$ & R$_\mathsf{T}$ & F1 & P$_\mathsf{T}$ & R$_\mathsf{T}$ \\
\midrule
Gemma-4 26B & 0.723 & 0.436 & \textbf{0.720} & \textbf{0.732} & \textbf{0.486} & 0.614 & 0.725 & 0.450 & 0.675 & \textbf{0.740} & \textbf{0.502} & 0.624 & 0.717 & 0.431 & \textbf{0.702} \\
Gemma-4 E4B & 0.571 & 0.245 & \textbf{0.802} & \textbf{0.677} & \textbf{0.397} & 0.520 & 0.628 & 0.322 & 0.665 & \textbf{0.641} & \textbf{0.339} & \textbf{0.681} & 0.603 & 0.302 & 0.636 \\
Qwen3.5 9B & 0.630 & 0.329 & \textbf{0.681} & \textbf{0.641} & \textbf{0.526} & 0.316 & 0.633 & 0.409 & 0.523 & \textbf{0.656} & \textbf{0.568} & 0.343 & 0.617 & 0.307 & \textbf{0.629} \\
EuroLLM 22B & 0.451 & 0.188 & \textbf{0.854} & \textbf{0.657} & \textbf{0.451} & 0.430 & 0.554 & 0.289 & 0.653 & \textbf{0.601} & \textbf{0.422} & 0.522 & 0.507 & 0.248 & \textbf{0.751} \\
\bottomrule
\end{tabular}}
\caption{$F1_{\text{macro}}$ and true-class precision (P$_\mathsf{T}$) and recall (R$_\mathsf{T}$) marginalised over each factor of the balance $\times$ order ablation (ratio levels averaged over order; order levels averaged over ratio). \textbf{Bold}: best value per model within each factor block.}
\label{tab:ablation-pr-marginal}
\end{table*}

% Detailed $F1_{\text{macro}}$ scores for all ratio-ordering ablations are reported in Table~\ref{tab:ablation-balance-order}. The \emph{true-first} approach in terms of demonstration ordering performs best for all models except \mbox{EuroLLM 22B}, for which \emph{random} ordering is better. For the \mbox{Qwen3.5 9B} model, \emph{balanced} demonstration ratio is best, while for all other models, a demonstration ratio \emph{proportional} to the training dataset performs best.

 Detailed $F1_{\text{macro}}$ scores for all ratio--ordering combinations are reported in Table~\ref{tab:ablation-pr-grid}. Per-factor mean $F1_{\text{macro}}$ scores (averaged over the other factor) are reported in Table~\ref{tab:ablation-pr-marginal}.
 In terms of per-factor mean $F1_{\text{macro}}$, the pattern is uniform: a \emph{proportional} class ratio outperforms \emph{balanced} demonstrations for all four models, and \emph{true-first} ordering outperforms both \emph{random} and \emph{true-last} for all four models, with \emph{true-last} consistently worst. The best individual combination deviates from these main effects for two models---\mbox{Qwen3.5 9B} peaks at \emph{balanced}~/~\emph{true-first} ($0.704$) and \mbox{EuroLLM 22B} at \emph{proportional}~/~\emph{random} ($0.663$)---reflecting the ratio--order interaction discussed in Section~\ref{ablation} (Fig.~\ref{fig:ablation_heat}).

 However, True-class Precision and Recall in Table~\ref{tab:ablation-pr-marginal} show another picture. Recall peaks for \textit{balanced} and \textit{true-last} demonstration sets for most models, while Precision is best for \emph{proportional} and \emph{true-first} demonstration sets across all models. As we argue that Recall is the most important metric for this task (while keeping an eye on Precision), \emph{balanced}, \emph{true-last} configuration for ratio~/~ordering of demonstrations comes out on top.

 \subsection{Shared Task Results}\label{app:shared-task}

We submitted five runs to the DEF Subtask; the organisers scored them against the held-out test labels (Table~\ref{tab:shared-task}). Our best submission, the fine-tuned \mbox{Gemma-4 E4B (FT)} (MUCnoHARM2), reached $0.74$, narrowly ahead of the prompted \mbox{Gemma-4 26B} (MUCnoHARM1, $0.72$). These held-out scores closely track the cross-validated results of Section~\ref{sec:res} ($0.741 \rightarrow 0.74$ for the fine-tune and $0.733 \rightarrow 0.72$ for \mbox{Gemma-4 26B}), indicating good transferability to the official test set. For the same \mbox{Gemma-4 26B} configuration, the default \emph{balanced}, \emph{randomly} ordered demonstrations (MUCnoHARM1) slightly outperformed the \emph{proportional}, \emph{true-first} variant (MUCnoHARM5, $0.70$), suggesting that the gains from tuning demonstration ratio and ordering did not transfer to the held-out data.

\begin{table}[ht]
\centering
\small
\resizebox{.45\textwidth}{!}{
\begin{tabular}{llllr}
\toprule
Run & Model & Cond. & Demonstrations & $F1_{\text{macro}}$ \\
\midrule
2 & G E4B (FT) & Impl. & zero-shot & $\mathbf{0.74}$ \\
1 & G 26B & Impl. & D-Sim, bal.~/~rand. & $0.72$ \\
5 & G 26B & Impl. & D-Sim, prop.~/~T-first & $0.70$ \\
4 & Q 9B & Impl. & Rand, bal.~/~T-first & $0.64$ \\
3 & G E4B & Desc. & D-Div, prop.~/~T-first & $0.56$ \\
\bottomrule
\end{tabular}}
\caption{Official shared task $F1_{\text{macro}}$ on the held-out DEF test set. Demonstrations: Dense-Similarity (\textit{D-Sim.}), Random (\textit{Rand}), Dense-Diversity (\textit{D-Div}) retrieval configurations, balanced (\textit{bal.}) or proportional (\textit{prop.}) class ratio~/~random (\textit{rand.}), true-first (\textit{T-first}) or true-last (\textit{T-last}) ordering. Best result in \textbf{bold}.}\label{tab:shared-task}
\end{table}

%\clearpage
\onecolumn
\begin{longtable}{lllrrrr}
\caption{Full results for all 112 configurations + GPT and fine-tuned models, sorted by $F1_{\text{macro}}$ (descending). P$_\mathsf{T}$~/~R$_\mathsf{T}$: Precision~/~Recall for the criminal class. FS~/~ZS: few-shot~/~zero-shot. D~/~S~/~F: dense~/~sparse~/~fusion embeddings. Sim~/~Div~/~MMR~/~Rand: Similarity~/~Diversity~/~MMR~/~Random Retrieval.}\label{tab:full}\\
\toprule
Model & Prompt & Retrieval & F1 & Acc. & P$_\mathsf{T}$ & R$_\mathsf{T}$ \\
\midrule
\endfirsthead
\multicolumn{7}{l}{\small\emph{(continued)}} \\
\toprule
Model & Prompt & Retrieval & F1 & Acc. & P$_\mathsf{T}$ & R$_\mathsf{T}$ \\
\midrule
\endhead
\midrule
\multicolumn{7}{r}{\small\emph{(continued on next page)}} \\
\endfoot
\bottomrule
\endlastfoot
Gemma-4 E4B (FT) & Implicit & ZS & 0.741 & 0.904 & 0.702 & 0.431 \\
Gemma-4 26B & Implicit & FS, D-Sim & 0.733 & 0.850 & 0.445 & 0.740 \\
Gemma-4 26B & Implicit & FS, D-MMR & 0.723 & 0.841 & 0.428 & 0.742 \\
Gemma-4 26B & Implicit & FS, D-Div & 0.722 & 0.833 & 0.417 & 0.788 \\
Gemma-4 26B & Implicit & FS, F-Sim & 0.719 & 0.836 & 0.419 & 0.745 \\
Gemma-4 26B & Implicit & FS, Rand & 0.717 & 0.829 & 0.410 & 0.790 \\
Gemma-4 26B & Description & FS, D-Sim & 0.716 & 0.830 & 0.411 & 0.773 \\
GPT-5.5 & Explicit & FS, D-Sim & 0.716 & 0.864 & 0.415 & 0.747 \\
Gemma-4 26B & Implicit & FS, S-Sim & 0.714 & 0.830 & 0.409 & 0.761 \\
Gemma-4 26B & Explicit & FS, D-MMR & 0.711 & 0.859 & 0.457 & 0.564 \\
Gemma-4 26B & Description & FS, F-Sim & 0.709 & 0.820 & 0.396 & 0.793 \\
Gemma-4 26B & Description & FS, D-MMR & 0.709 & 0.821 & 0.397 & 0.785 \\
Gemma-4 26B & Explicit & FS, F-Sim & 0.701 & 0.849 & 0.430 & 0.571 \\
Gemma-4 26B & Explicit & FS, D-Sim & 0.698 & 0.859 & 0.451 & 0.506 \\
Gemma-4 26B & Description & FS, S-Sim & 0.697 & 0.807 & 0.378 & 0.800 \\
Gemma-4 26B & Description & FS, D-Div & 0.696 & 0.799 & 0.373 & 0.848 \\
Gemma-4 26B & Title & FS, D-Div & 0.695 & 0.808 & 0.378 & 0.783 \\
Gemma-4 26B & Description & FS, Rand & 0.695 & 0.799 & 0.372 & 0.846 \\
Gemma-4 26B & Title & FS, D-MMR & 0.693 & 0.811 & 0.377 & 0.749 \\
Gemma-4 26B & Title & FS, Rand & 0.692 & 0.805 & 0.373 & 0.781 \\
Gemma-4 26B & Title & FS, D-Sim & 0.691 & 0.815 & 0.380 & 0.713 \\
Gemma-4 26B & Explicit & FS, D-Div & 0.691 & 0.847 & 0.421 & 0.532 \\
Gemma-4 26B & Title & FS, F-Sim & 0.690 & 0.809 & 0.374 & 0.747 \\
Gemma-4 26B & Title & FS, S-Sim & 0.686 & 0.803 & 0.366 & 0.757 \\
Gemma-4 26B & Explicit & FS, Rand & 0.681 & 0.848 & 0.416 & 0.492 \\
Gemma-4 26B & Explicit & FS, S-Sim & 0.676 & 0.836 & 0.392 & 0.523 \\
Gemma-4 26B & Implicit & ZS & 0.676 & 0.791 & 0.353 & 0.764 \\
Gemma-4 26B & Description & ZS & 0.674 & 0.782 & 0.347 & 0.807 \\
Gemma-4 E4B & Explicit & FS, S-Sim & 0.659 & 0.799 & 0.341 & 0.622 \\
Gemma-4 26B & Explicit & ZS & 0.652 & 0.777 & 0.329 & 0.683 \\
Gemma-4 E4B & Explicit & FS, D-Sim & 0.650 & 0.781 & 0.324 & 0.663 \\
Gemma-4 E4B & Explicit & FS, Rand & 0.646 & 0.798 & 0.328 & 0.559 \\
Gemma-4 E4B & Explicit & FS, F-Sim & 0.644 & 0.777 & 0.317 & 0.653 \\
Gemma-4 E4B & Explicit & FS, D-MMR & 0.644 & 0.762 & 0.313 & 0.733 \\
Gemma-4 E4B & Explicit & FS, D-Div & 0.640 & 0.792 & 0.319 & 0.559 \\
Qwen3.5 9B & Explicit & FS, D-Sim & 0.636 & 0.821 & 0.338 & 0.426 \\
Gemma-4 26B & Title & ZS & 0.630 & 0.758 & 0.299 & 0.670 \\
Qwen3.5 9B & Implicit & FS, Rand & 0.630 & 0.740 & 0.298 & 0.769 \\
Qwen3.5 9B & Explicit & FS, D-MMR & 0.628 & 0.818 & 0.326 & 0.407 \\
Qwen3.5 9B & Explicit & FS, F-Sim & 0.627 & 0.818 & 0.326 & 0.402 \\
Qwen3.5 9B & Description & FS, D-Div & 0.626 & 0.752 & 0.294 & 0.677 \\
Qwen3.5 9B & Description & FS, Rand & 0.626 & 0.749 & 0.293 & 0.689 \\
Gemma-4 E4B & Description & FS, D-Div & 0.612 & 0.715 & 0.280 & 0.788 \\
Qwen3.5 9B & Implicit & FS, D-Div & 0.610 & 0.722 & 0.277 & 0.733 \\
Gemma-4 E4B & Description & FS, Rand & 0.609 & 0.707 & 0.277 & 0.807 \\
Qwen3.5 9B & Explicit & FS, S-Sim & 0.606 & 0.839 & 0.336 & 0.275 \\
Gemma-4 E4B & Title & FS, D-Div & 0.600 & 0.706 & 0.267 & 0.747 \\
Gemma-4 E4B & Title & FS, D-Sim & 0.600 & 0.706 & 0.267 & 0.749 \\
Qwen3.5 9B & Explicit & ZS & 0.598 & 0.811 & 0.286 & 0.325 \\
Gemma-4 E4B & Description & FS, D-Sim & 0.598 & 0.697 & 0.266 & 0.788 \\
Qwen3.5 9B & Implicit & FS, F-Sim & 0.597 & 0.705 & 0.264 & 0.737 \\
Qwen3.5 9B & Implicit & FS, D-MMR & 0.597 & 0.702 & 0.265 & 0.754 \\
Qwen3.5 9B & Implicit & FS, S-Sim & 0.597 & 0.708 & 0.263 & 0.723 \\
Gemma-4 E4B & Description & ZS & 0.597 & 0.706 & 0.263 & 0.733 \\
Qwen3.5 9B & Implicit & FS, D-Sim & 0.597 & 0.709 & 0.263 & 0.711 \\
Gemma-4 E4B & Title & FS, F-Sim & 0.597 & 0.695 & 0.265 & 0.793 \\
Gemma-4 E4B & Description & FS, F-Sim & 0.594 & 0.685 & 0.265 & 0.831 \\
Qwen3.5 9B & Description & FS, S-Sim & 0.593 & 0.715 & 0.258 & 0.660 \\
Gemma-4 E4B & Explicit & ZS & 0.588 & 0.795 & 0.264 & 0.340 \\
Gemma-4 E4B & Title & FS, S-Sim & 0.582 & 0.686 & 0.250 & 0.735 \\
Gemma-4 E4B & Implicit & FS, D-Sim & 0.582 & 0.674 & 0.253 & 0.800 \\
Qwen3.5 9B & Description & FS, F-Sim & 0.580 & 0.692 & 0.246 & 0.692 \\
Qwen3.5 9B & Title & FS, D-Div & 0.578 & 0.689 & 0.245 & 0.692 \\
Gemma-4 E4B & Title & FS, D-MMR & 0.577 & 0.667 & 0.251 & 0.815 \\
Qwen3.5 9B & Description & FS, D-Sim & 0.576 & 0.683 & 0.244 & 0.711 \\
Gemma-4 E4B & Description & FS, S-Sim & 0.575 & 0.667 & 0.248 & 0.795 \\
Gemma-4 E4B & Title & FS, Rand & 0.574 & 0.680 & 0.242 & 0.711 \\
Qwen3.5 9B & Description & FS, D-MMR & 0.573 & 0.675 & 0.243 & 0.733 \\
Gemma-4 E4B & Implicit & FS, Rand & 0.572 & 0.663 & 0.245 & 0.793 \\
Gemma-4 E4B & Implicit & FS, D-Div & 0.572 & 0.663 & 0.245 & 0.793 \\
Qwen3.5 9B & Explicit & FS, Rand & 0.570 & 0.853 & 0.340 & 0.164 \\
Qwen3.5 9B & Explicit & FS, D-Div & 0.568 & 0.849 & 0.321 & 0.166 \\
Gemma-4 E4B & Description & FS, D-MMR & 0.567 & 0.646 & 0.245 & 0.860 \\
Gemma-4 E4B & Implicit & FS, S-Sim & 0.561 & 0.646 & 0.238 & 0.815 \\
Qwen3.5 9B & Implicit & ZS & 0.558 & 0.873 & 0.500 & 0.113 \\
Gemma-4 E4B & Implicit & FS, F-Sim & 0.557 & 0.639 & 0.235 & 0.819 \\
Qwen3.5 9B & Title & FS, Rand & 0.553 & 0.668 & 0.220 & 0.634 \\
Gemma-4 E4B & Implicit & FS, D-MMR & 0.548 & 0.622 & 0.232 & 0.855 \\
Qwen3.5 9B & Description & ZS & 0.541 & 0.660 & 0.207 & 0.593 \\
Qwen3.5 9B & Title & FS, S-Sim & 0.538 & 0.633 & 0.214 & 0.704 \\
Gemma-4 E4B & Title & ZS & 0.525 & 0.606 & 0.210 & 0.759 \\
Qwen3.5 9B & Title & FS, F-Sim & 0.511 & 0.586 & 0.202 & 0.764 \\
Qwen3.5 9B & Title & ZS & 0.509 & 0.614 & 0.184 & 0.595 \\
Qwen3.5 9B & Title & FS, D-Sim & 0.507 & 0.580 & 0.201 & 0.771 \\
Qwen3.5 9B & Title & FS, D-MMR & 0.507 & 0.576 & 0.202 & 0.790 \\
EuroLLM 22B & Explicit & FS, S-Sim & 0.474 & 0.539 & 0.180 & 0.740 \\
EuroLLM 22B & Explicit & FS, D-Sim & 0.473 & 0.523 & 0.190 & 0.843 \\
EuroLLM 22B & Explicit & FS, F-Sim & 0.464 & 0.517 & 0.182 & 0.798 \\
EuroLLM 22B & Implicit & FS, D-Sim & 0.447 & 0.485 & 0.182 & 0.872 \\
GPT-5.5 & Title & ZS & 0.440 & 0.464 & 0.192 & 1.000 \\
EuroLLM 22B & Explicit & FS, D-MMR & 0.438 & 0.476 & 0.176 & 0.851 \\
EuroLLM 22B & Description & FS, D-Sim & 0.436 & 0.474 & 0.175 & 0.843 \\
EuroLLM 22B & Explicit & FS, Rand & 0.434 & 0.479 & 0.166 & 0.769 \\
EuroLLM 22B & Implicit & FS, F-Sim & 0.433 & 0.467 & 0.177 & 0.875 \\
EuroLLM 22B & Description & FS, S-Sim & 0.428 & 0.465 & 0.170 & 0.829 \\
EuroLLM 22B & Implicit & FS, S-Sim & 0.427 & 0.459 & 0.174 & 0.870 \\
EuroLLM 22B & Description & FS, F-Sim & 0.425 & 0.459 & 0.172 & 0.848 \\
Gemma-4 E4B & Implicit & ZS & 0.420 & 0.448 & 0.174 & 0.892 \\
EuroLLM 22B & Explicit & FS, D-Div & 0.417 & 0.453 & 0.164 & 0.805 \\
EuroLLM 22B & Implicit & FS, D-MMR & 0.416 & 0.443 & 0.173 & 0.896 \\
EuroLLM 22B & Description & FS, Rand & 0.404 & 0.431 & 0.166 & 0.863 \\
EuroLLM 22B & Description & FS, D-MMR & 0.399 & 0.422 & 0.166 & 0.884 \\
EuroLLM 22B & Title & FS, D-Sim & 0.395 & 0.418 & 0.165 & 0.879 \\
EuroLLM 22B & Title & FS, S-Sim & 0.392 & 0.413 & 0.165 & 0.889 \\
EuroLLM 22B & Description & FS, D-Div & 0.389 & 0.413 & 0.158 & 0.834 \\
EuroLLM 22B & Title & FS, F-Sim & 0.385 & 0.403 & 0.165 & 0.908 \\
EuroLLM 22B & Implicit & FS, D-Div & 0.367 & 0.382 & 0.159 & 0.899 \\
EuroLLM 22B & Title & FS, D-Div & 0.362 & 0.379 & 0.151 & 0.841 \\
EuroLLM 22B & Title & FS, Rand & 0.358 & 0.376 & 0.149 & 0.829 \\
EuroLLM 22B & Explicit & ZS & 0.358 & 0.366 & 0.166 & 0.990 \\
EuroLLM 22B & Implicit & FS, Rand & 0.356 & 0.370 & 0.153 & 0.872 \\
EuroLLM 22B & Title & FS, D-MMR & 0.352 & 0.363 & 0.157 & 0.918 \\
EuroLLM 22B & Description & ZS & 0.222 & 0.223 & 0.137 & 0.969 \\
EuroLLM 22B & Implicit & ZS & 0.211 & 0.212 & 0.135 & 0.964 \\
EuroLLM 22B & Title & ZS & 0.169 & 0.174 & 0.133 & 0.990 \\
\end{longtable}

\end{document}